\documentclass[letterpaper]{article} % DO NOT CHANGE THIS
\usepackage{aaai2026}  % DO NOT CHANGE THIS
\nocopyright % Do not display copyright
\usepackage{times}  % DO NOT CHANGE THIS
\usepackage{helvet}  % DO NOT CHANGE THIS
\usepackage{courier}  % DO NOT CHANGE THIS
\usepackage[hyphens]{url}  % DO NOT CHANGE THIS
\usepackage{graphicx} % DO NOT CHANGE THIS
\usepackage{natbib}  % DO NOT CHANGE THIS AND DO NOT ADD ANY OPTIONS TO IT
\usepackage{caption} % DO NOT CHANGE THIS AND DO NOT ADD ANY OPTIONS TO IT
\usepackage{algorithm}
\usepackage{algorithmic}

\usepackage{newfloat}
\usepackage{listings}
\DeclareCaptionStyle{ruled}{labelfont=normalfont,labelsep=colon,strut=off} % DO NOT CHANGE THIS
\floatstyle{ruled}
\newfloat{listing}{tb}{lst}{}
\floatname{listing}{Listing}
\PassOptionsToPackage{table}{xcolor}
\usepackage{colortbl}
\usepackage{tcolorbox}
\tcbuselibrary{skins,breakable}
\usepackage{booktabs}
\usepackage{multirow}
\usepackage{makecell}
\usepackage{graphicx} 
\usepackage{subcaption}
\usepackage{makecell}

\title{On the Alignment of Large Language Models with Global Human Opinion}
\author {
    Yang Liu\textsuperscript{\rm 1},
    Masahiro Kaneko\textsuperscript{\rm 2},
    Chenhui Chu\textsuperscript{\rm 1}
}
\affiliations {
    \textsuperscript{\rm 1}Kyoto University\\
    \textsuperscript{\rm 2}MBZUAI\\
    yangliu@nlp.ist.i.kyoto-u.ac.jp, Masahiro.Kaneko@mbzuai.ac.ae, chu@i.kyoto-u.ac.jp
}

\usepackage{bibentry}
\begin{document}

\maketitle

\begin{abstract}
Today's large language models (LLMs) are capable of supporting multilingual scenarios, allowing users to interact with LLMs in their native languages. When LLMs respond to subjective questions posed by users, they are expected to align with the views of specific demographic groups or historical periods, shaped by the language in which the user interacts with the model.
Existing studies mainly focus on researching the opinions represented by LLMs among demographic groups in the United States or a few countries, lacking worldwide country samples and studies on human opinions in different historical periods, as well as lacking discussion on using language to steer LLMs.
Moreover, they also overlook the potential influence of prompt language on the alignment of LLMs' opinions.
In this study, our goal is to fill these gaps.
To this end, we create an evaluation framework based on the World Values Survey (WVS) to systematically assess the alignment of LLMs with human opinions across different countries, languages, and historical periods around the world. 
We find that LLMs appropriately or over-align the opinions with only a few countries while under-aligning the opinions with most countries.
Furthermore, changing the language of the prompt to match the language used in the questionnaire can effectively steer LLMs to align with the opinions of the corresponding country more effectively than existing steering methods. At the same time, LLMs are more aligned with the opinions of the contemporary population.
To our knowledge, our study is the first comprehensive investigation of the topic of opinion alignment in LLMs across global, language, and temporal dimensions. 
Our code and data are publicly available at \url{https://github.com/ku-nlp/global-opinion-alignment} and \url{https://github.com/nlply/global-opinion-alignment}.

\end{abstract}

\section{Introduction}

Large language models (LLMs) have become crucial to everyday decision-making and assistance~\cite{openai2023gpt, bubeck2023sparks, bommasani2021opportunities}.
Once LLMs are deployed as products, they are inevitably asked to answer subjective questions, not just objective ones~\cite{ouyang2022training, santurkar2023whose,meister-etal-2025-benchmarking}.
From an alignment perspective, next-token prediction during pretraining gives LLMs a statistical prior over human opinions~\cite{radford2019language,brown2020language}, while subsequent alignment stages, such as instruction tuning~\cite{wei2021finetuned,sanh2021multitask} and preference optimization~\cite{christiano2017deep,ouyang2022training}, shape this prior toward the opinions in their preference datasets, which are typically annotated by annotators from various countries, speaking diverse languages, and across different ages~\cite{mason2012conducting}.

\begin{figure}[t]
\centering
\includegraphics[width=\columnwidth]{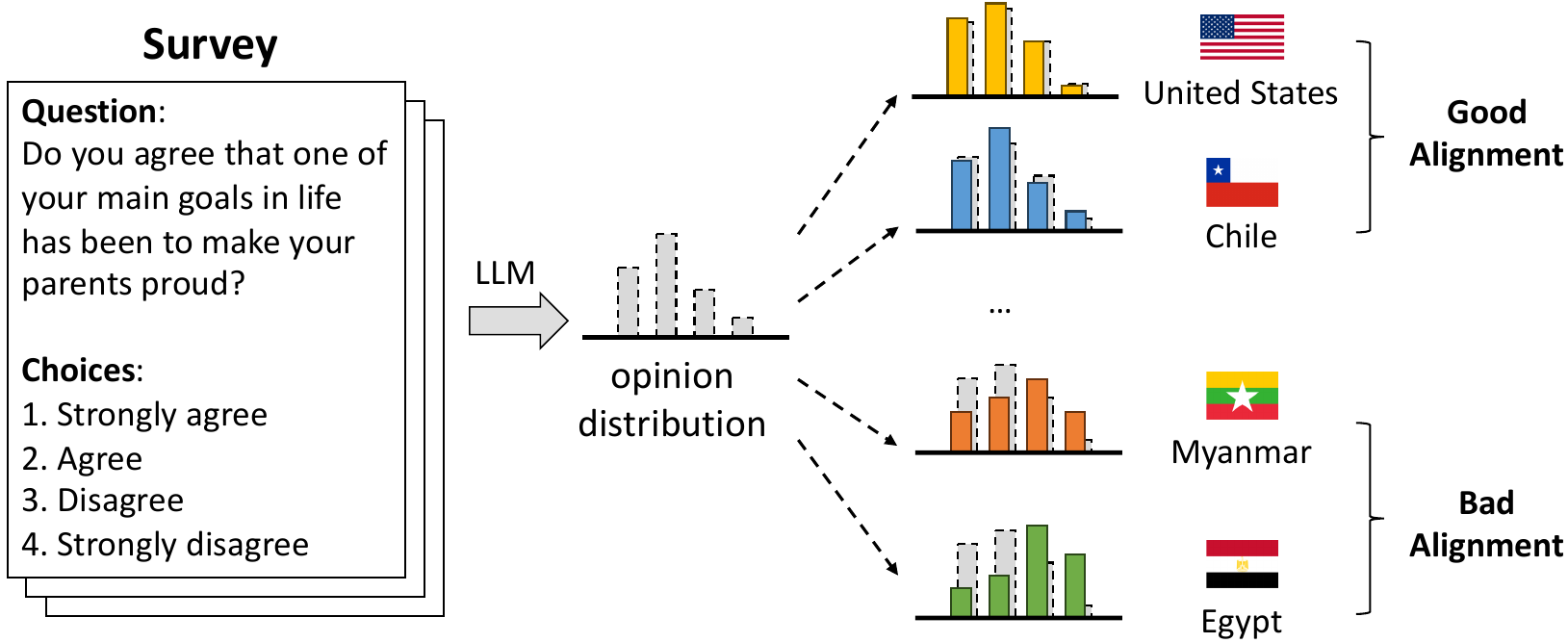}
\caption{An example of our main idea. When asking the LLM a subjective question, is the LLM's opinion distribution closer to specific countries' opinion distributions?}
\label{fig:demo}
\end{figure}

Existing study mainly focuses on exploring opinion alignment between LLMs and demographic groups in the United States~\cite{santurkar2023whose}.
\citet{durmus2023towards} expand~\citet{santurkar2023whose}'s study to explore the alignment between the LLM Claude and other countries, and they further find that translating the questions into Russian, Chinese, and Turkish fails to improve the alignment between the LLM and the language speakers.
A crucial challenge to previous studies is that the LLMs' opinion distribution is expressed as the next token log probabilities from the model, which suffers from a mis-calibration issue and requires the LLMs to be able to output logits~\cite{meister-etal-2025-benchmarking}.
Therefore, in this study, we adopt the verbalized distribution method to investigate the alignment between LLMs and human opinions. We expand to investigate seven state-of-the-art LLMs and eight languages covering 10 countries to validate the effectiveness of language steering. 
In addition, we innovatively explore the alignment of LLMs and human opinions in the temporal dimension.
Therefore, our study covers the alignment of LLMs with human opinions in three dimensions: \textbf{global, languages, and temporal}.

The key question we focus on is illustrated in Figure~\ref{fig:demo}: when asking the LLMs a subjective question, is the LLM's opinion distribution closer to specific countries' opinion distributions?
If an LLM systematically reflects the opinions of specific countries, this may lead to unexpected country dominance or misalignment in international applications~\cite{bender2021dangers, alkhamissi-etal-2024-investigating, sukiennik2025evaluation}.
Additionally, analyzing the opinions reflected by the LLMs can help improve the transparency and accountability of the LLMs, allowing developers to better assess and mitigate potential geopolitical or ethical risks. Finally, this study can provide a foundation for building more human opinion adaptive or neutral AI systems that respect diverse views in the globally interconnected world.

The World Values Survey~\citep[WVS;][]{wvs_round7_v6}, a comprehensive global research program, systematically collects human opinions on a wide range of topics, including politics, religion, and ethics.
The standardized questionnaire design of the WVS allows for consistent comparisons of opinions across countries, languages, and historical periods.
In addition, the survey covers highly subjective and value-oriented questions, which align well with the types of prompts used to probe LLMs' opinions.
By using WVS as the benchmark, in this study, we focus on the following research questions: 
\textbf{RQ1}: Do LLMs appropriately align with the opinions of different countries?
\textbf{RQ2}: Can language steer LLMs to express opinions closer to language speakers?
\textbf{RQ3}: Which human historical period's opinions are reflected in LLMs' responses?

Our work makes four main contributions: 
\textit{1)} We systematically investigate human opinion alignment in global, language, and temporal dimensions with LLMs;
\textit{2)} We reveal that current LLMs tend to align with the opinions of English- or Spanish-speaking countries such as the United States, Canada, and Chile, but under-align with the opinions of information-regulated or non-Latin-speaking countries such as Myanmar and Egypt;
\textit{3)} We employ language steering on eight languages, which is effective in over 82\% of the 240 cases (3 baselines\footnote{The three baselines include no steering, persona steering, and few-shot steering.} $\times$ 8 models $\times$ 10 countries) in our experiments;
\textit{4)} We evaluate the alignment of LLMs with human opinions in the temporal dimension, revealing that LLMs are most aligned with contemporary human opinions.

\section{The World Values Survey}
\label{sec:the_world_values_survey}
In this section, we explain why WVS is suitable for our study from three dimensions: global, language, and temporal.
Further details are provided in Appendix~\ref{sec:more_detals_of_wvs}.

\paragraph{Global Dimension} 
The WVS covers highly subjective and value-oriented questions, which align well with the types of prompts used to probe LLMs' opinions.
The latest wave covers survey data from 66 countries and regions, enabling us to compare the alignment of LLMs with different countries' human opinions.
In addition, each country has more than 1,000 human questionnaire results, which ensures that the survey data reflects real human behavior to some extent.
These features make it possible for us to study RQ1.

\paragraph{Language Dimension}
The WVS questionnaire was translated into all languages which serve as the first language for 15\% (or more) of the population.
For countries that participated in previous waves of the WVS, translation for the replicated items is used from the previous wave questionnaires to minimize bias during overtime comparison of answer distributions.\footnote{\url{https://www.worldvaluessurvey.org/wvs.jsp}}
According to our statistics, the latest wave provided more than 50 language translations, ensuring that the questionnaire covers languages supported by LLMs and providing linguistic diversity for RQ2.

\paragraph{Temporal Dimension}
Since 1981, the WVS has conducted seven waves of globally representative surveys.
The WVS maintains functional consistency across most questions in all waves, allowing us to study long-term trends of alignment between LLMs and human opinions across different waves, providing temporal data support for our RQ3.

\section{Experiment Settings}
In this section, we introduce the experiment settings. 
More details such as the question filter rules and the complete country list (\S\ref{subsec:dataset}), the full prompts (\S\ref{subsec:prompts}), and more details of the distribution expression methods (\S\ref{subsec:distribution_expression_methods}) are provided in Appendix~\ref{sec:experiment_settings}.

\subsection{Dataset}
\label{subsec:dataset}
In RQ1, we conduct a question filter process on the English version of wave 7, which includes survey data from 66 countries, to filter out questions that are not suitable for this study. For example, questions that require actual experience, objectivity, etc. After this filter process, we get 144 opinion-related questions. 
In RQ2, we used the same 144 questions and collected them from other language versions of the questionnaire.
In RQ3, to ensure a certain number of common questions, we chose to analyze WVS's data from wave 5, wave 6, and wave 7.\footnote{Our statistical result indicates that, when wave 4 is taken into account, there are only 48 common questions.}
Each wave corresponds to a historical questionnaire version. 
We organized each English version from wave 5 to 7 and then selected questions that are common in all three waves. We obtained 75 opinion-related questions that are shared across all three waves.

\subsection{Models}
We experiment with seven instruction-tuned multilingual LLMs: Aya-23-35B~\cite{aryabumi2024aya}, Llama3-70B-Instruct~\cite{llama3modelcard}, Qwen2.5-72B-Instruct~\cite{qwen2.5}, GPT-3.5-Turbo~\citep[\texttt{gpt-3.5-turbo-0125};][]{openai_gpt35turbo_docs}, GPT-4~\citep[\texttt{gpt-4-0613};][]{achiam2023gpt}, GPT-5~\citep[\texttt{gpt-5};][]{openai2025gpt5}, DeepSeek-V3~\citep[\texttt{DeepSeek-V3-0324};][]{liu2024deepseek}, and DeepSeek-R1~\citep[\texttt{DeepSeek-R1-0528};][]{guo2025deepseek}.
For open-weight community models, we select the largest publicly released weights in each series. 
For commercial LLMs, we include OpenAI's GPT-3.5-Turbo, GPT-4, and GPT-5 (closed-source models), as well as DeepSeek-V3 and DeepSeek-R1, noting that DeepSeek releases open-weight models under a permissive license.

\subsection{Prompts}
\label{subsec:prompts}
As the instruction fine-tuned LLMs use ``task instructions + examples'' as input during training, it is easier for them to follow the requirements of the prompt and output content in the required format.
We refer to~\citet{meister-etal-2025-benchmarking} to control the LLMs' output distribution in JSON format (e.g., \texttt{\{1: 31\%, 2: 4\%, 3: 30\%, 4: 35\%\}}) by providing five few-shot examples.

\subsection{Distribution Expression Methods} 
\label{subsec:distribution_expression_methods}
\paragraph{LLMs' opinion distribution}
Being able to accurately express LLMs' opinion distributions is a precondition of our study.
\citet{meister-etal-2025-benchmarking}'s work compares three distribution expression methods: \textit{model log-probabilities}, \textit{sequence of tokens}, and \textit{verbalized distribution}.
Their analysis reveals that \textit{verbalized distribution} outperforms the other two methods.
Therefore, we use \textit{verbalized distribution} to represent the opinion distribution $D_{\mathcal{M}}(q)$ of the LLM $\mathcal{M}$ on answering question $q$.

\paragraph{Human opinion distribution}
We calculate human opinion distributions from the statistical data of the WVS, which contain more than 1,000 pieces of human data for each surveyed country.
The distribution of the country $c \in \mathcal{C}$ answers to the question $q \in \mathcal{Q}$ with $|\mathcal{N}|$ options can be denoted as $D_c(q) = \{D_{c,n} | n \in \mathcal{N}\}$. 
Specifically, $D_{c,n}$ is as follows:
\begin{equation}
    D_{c,n} = \frac{|\mathcal{P}^{(n)}_{c}|}{\sum_{i \in \mathcal{N}} |\mathcal{P}^{(i)}_{c}|}
\end{equation}
where $|\mathcal{P}_c^{(n)}|$ denotes the number of people in the country $c$ who choose option $n$ when answering question $q$.

\subsection{Alignment Metrics}
Measuring the alignment between the LLM's opinion distribution $D_{\mathcal{M}}(q)$ and human opinion distribution $D_{c}(q)$ requires considering the order of the options in these two distributions. Moreover, it is desirable for the metrics to range from 0 to 1. In this paper, we use the \textit{alignment} proposed by~\citet{santurkar2023whose} as our metric. This metric is used to measure the alignment score between distributions $D_{\mathcal{M}}$ and $D_{c}$. The formal definition of \textit{alignment} is as follows:
\begin{equation}
    \mathcal{A}(D_{\mathcal{M}},D_{c};\mathcal{Q})
= \frac{1}{|\mathcal{Q}|}
  \sum_{q \in \mathcal{Q}}
  \left(
    1 - \frac{\mathrm{WD} (D_{\mathcal{M}}(q), D_{c}(q) )}{|\mathcal{N}|-1}
  \right)
\end{equation}
where $\mathcal{Q}$ is the question set and $|\mathcal{N}|$ is the number of options of the question $q$. The denominator $|\mathcal{N}|-1$ serves to normalize the Wasserstein distance ($\mathrm{WD}$)\footnote{\url{https://en.wikipedia.org/wiki/Wasserstein_metric}} so that the value of the metric falls between 0 and 1. The metric's value of 1 indicates that the two distributions are perfectly matched.

\section{Global Opinion Alignment (RQ1)}
\label{sec:global_opinion_representation}

\begin{figure*}[htbp]
    \centering
    % Overall alignment scores of LLMs
    \begin{subfigure}[b]{0.37\textwidth}
        \centering
        \includegraphics[width=\linewidth]{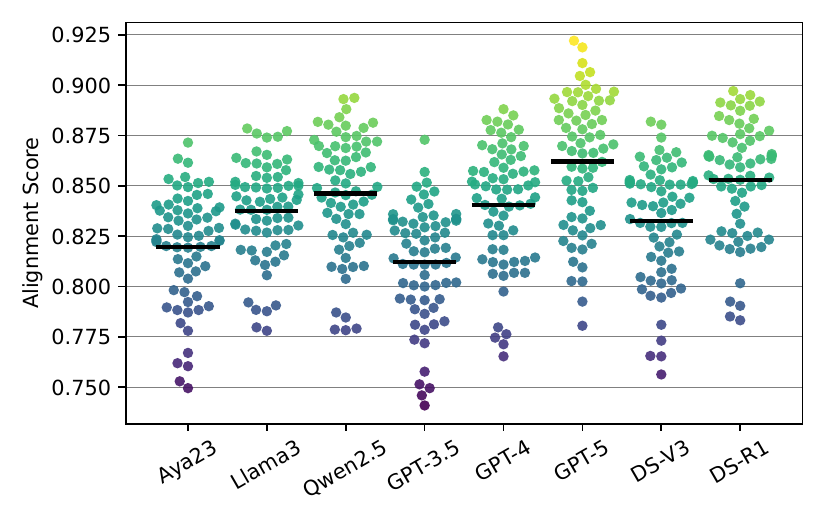}
        \caption{The alignment score of LLMs with each country.}
        \label{fig:rq1_sub1}
    \end{subfigure}
    \hfill
    % Top 6 and Bottom 6 countries
    \begin{subfigure}[b]{0.308\textwidth}
        \centering
        \includegraphics[width=\linewidth]{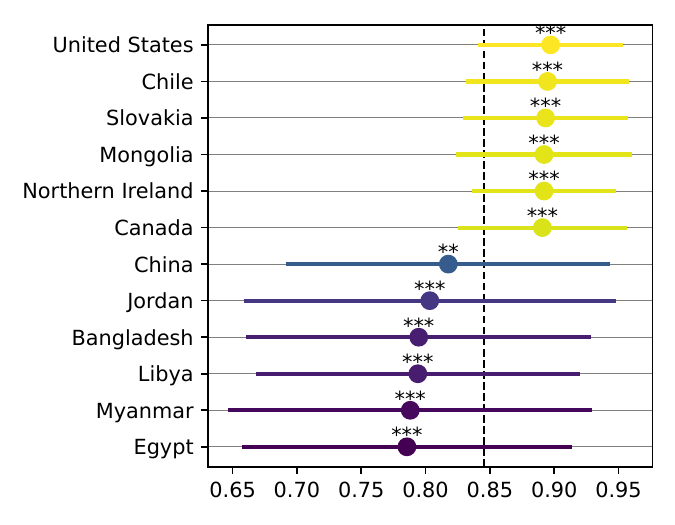}
        \caption{The first and last 6 countries or regions.}
        \label{fig:rq1_sub2}
    \end{subfigure}
    \begin{subfigure}[b]{0.308\textwidth}
        \centering
        \includegraphics[width=\linewidth]{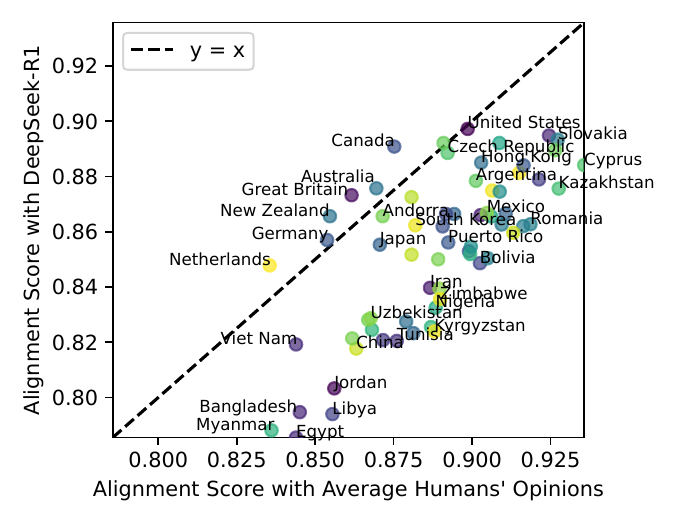}
        \caption{DeepSeek-R1 vs. Human}
        \label{fig:rq1_sub3}
    \end{subfigure}
    
    \caption{The overall results of \textbf{RQ1}. (a) The alignment score of LLMs with each country, where each point represents a country and the black line represents the average alignment score of all countries. ``DS'' is the abbreviation for ``DeepSeek.'' (b) The first and last 6 countries ranked by their alignment scores with DeepSeek-R1's opinion distributions, the point represents the average alignment score of the country on all questions and the line represents the standard deviation. *** denotes $p$-value $<$ 0.001 ($t$-test). (c) The relationship of different countries to DeepSeek-R1's opinion distribution and the average human opinion distribution alignment scores. For visibility, we hide some countries. See Appendix~\ref{sec:alignment_difference_level} for the full version.}
    \label{fig:rq1_whole}
\end{figure*}

In this section, we calculate the alignment score between LLMs' opinion distributions and human opinion distributions in wave 7. %, which covers 66 countries.
Then, we discuss the alignment of LLMs with human opinion at three levels: \textit{1)} \textbf{Model level}: Which LLM most matches the human opinion distribution? \textit{2)} \textbf{Country level}: Which countries' opinions are aligned by LLMs? \textit{3)} \textbf{Alignment difference level}: If we take the average human opinion distribution as a possible goal, how do LLMs and this average distribution differ in terms of alignment with a country?

\subsection{Model Level} 

As shown in Figure~\ref{fig:rq1_whole}(a), we demonstrate the overall alignment scores of LLMs' opinion distributions with human opinion distributions.
The experimental results show that GPT-5 achieves the highest average alignment score (0.8621). Meanwhile, the average alignment scores of Llama3, Qwen2.5, GPT-4, DeepSeek-V3, and DeepSeek-R1 closely follow those of GPT-5, also demonstrating high human opinion alignment performance.
In contrast, Aya-23-35B and GPT-3.5-Turbo, as earlier generations of instruction fine-tuning LLMs, have lower average alignment scores and a number of countries with low tail scores; the relative lack of multilingual long-tail coverage may be part of the reason.
In addition, the alignment scores of each LLM differ significantly across countries, indicating that no single LLM can simultaneously align with all countries' opinions.

\subsection{Country Level} 
Next, we move our attention to the question of \textit{which countries LLMs are more aligned with}.
% The results in the above section show that DeepSeek-R1 achieves the highest mean alignment scores with human opinion distributions.
Figure~\ref{fig:rq1_whole}(b) shows the five countries' opinion distributions that are most and least aligned with DeepSeek-R1' opinion distributions.\footnote{Results for all LLMs are available in Appendix~\ref{sec:more_results_for_country_level}, and the complete list of countries are provided in Appendix~\ref{sec:complete_country_list}.}
We can see that DeepSeek-R1 has the highest average alignment scores with countries such as the United States and Chile (0.8972 and 0.8948), and relatively low scores with countries such as Egypt and Myanmar (0.7855 and 0.7881). Notably, the standard deviation of the alignment scores in the high-alignment countries is significantly lower than that in the low-alignment countries, indicating that DeepSeek-R1 is both \textbf{approximate} and \textbf{robust} in the high-alignment countries. In contrast, low-alignment countries suffer from both systematic misalignment and high standard deviations.
This may be due to the widespread presence of English and Spanish content,\footnote{\url{https://en.wikipedia.org/wiki/Languages_used_on_the_Internet}} which makes the LLMs highly homogeneous with the dominant views of the United States, Canada, and Chile societies, while low-aligned countries tend to be in non-Latin-speaking environments, which results in a severe scarcity of their political and cultural alignments in the training set. 
In addition, the alignment with Slovakia and Mongolia is surprisingly high, which implies that the LLM may have learned indirect social signals to Slovakia and Mongolia by borrowing from regional media or cross-language interconnected corpora, suggesting that estimating cultural alignment purely in terms of corpus size is still insufficient. It is worth noting that although DeepSeek-R1 supports Chinese well, its opinion distribution is not highly aligned with China. In fact, as we show in Appendix~\ref{sec:more_results_of_rq1}, DeepSeek-R1 exhibits a similar pattern to other LLMs (such as GPT-4) in terms of country's opinion alignment.

\subsection{Alignment Difference Level} 
It is impossible to align LLMs with all countries at the same time due to the diversity of cultures. 
LLMs aligning with the average human opinion distribution is a plausible solution, and we hypothesize that it is the ideal goal. 
Then, we take the average human opinion distribution as the baseline to investigate how well LLMs fit the different countries' opinions.
Figure~\ref{fig:rq1_whole}(c) illustrates the relationship of different countries to the DeepSeek-R1's opinion distribution and the average human opinion distribution alignment scores. Consider $y=x$ as the reference line: points that fall on the line indicate that the LLM \textbf{appropriately align} with the country's opinions, points that fall above the line indicate that the LLM \textbf{over-align} with the country's opinions, and points that fall below the line indicate \textbf{under-alignment}. We can see that DeepSeek-R1 most appropriately aligns with the United States' opinions, as well as over-aligns with some developed countries' opinions.
It is worth noting that although DeepSeek-R1 is relatively close to Chile, Slovakia, and Mongolia's opinion distributions (Figure~\ref{fig:rq1_whole}(b)), when compared to the average human opinion distribution, DeepSeek-R1 under-aligns with these countries even more (Figure~\ref{fig:rq1_whole}(c)).
Overall, DeepSeek-R1 appropriately aligns or over-aligns with the views of a few countries, while under-aligning with the views of most countries, especially developing countries such as Myanmar, Egypt, and Libya.

\section{Steerability of Language (RQ2)}
\begin{table*}[t]
    \centering
    \scriptsize
    \begin{tabular}{llllllllll}
    \toprule
    \textbf{Method} & \textbf{Aya23} & \textbf{Llama3} & \textbf{Qwen2.5} & \textbf{GPT-3.5} & \textbf{GPT-4} & \textbf{GPT-5} & \textbf{DS-V3} & \textbf{DS-R1} & \textbf{AVG.} \\
    
    \midrule
    \rowcolor{gray!10} \multicolumn{10}{c}{ \textit{China}} \\
    \midrule
    
    No Steering (En.) & 0.8038 & 0.8235 & 0.8273 & 0.7816 & 0.8152 & 0.8348 & 0.7952 & 0.8174 & 0.8124 \\ 
    \ \ +\textit{Language Steering} (Zh.) & \textbf{0.8103} & \textbf{0.8308} & \textbf{0.8439} & \textbf{0.7976} & \textbf{0.8526}$^{*}$ & \textbf{0.8601} & \textbf{0.8471}$^{**}$ & \textbf{0.8498}$^{*}$ & \textbf{0.8365}$^{***}$ \\ 
    
    \hline 
    
    Persona Steering (En.)  & 0.8154 & 0.8439 & 0.8316 &0.7846 & 0.8247 & 0.8953$^{***}$ & 0.8278$^{*}$ & 0.8625$^{**}$ & 0.8357$^{***}$ \\ 
    \ \ +\textit{Language Steering} (Zh.) & \textbf{0.8216} & \textbf{0.8553}$^{*}$ & \textbf{0.8629}$^{*}$ & \textbf{0.8088} & \textbf{0.8718}$^{***}$ & \textbf{0.8979}$^{***}$ & \textbf{0.8732}$^{***}$ & \textbf{0.8828}$^{***}$ & \textbf{0.8593}$^{***}$ \\ 
    
    \hline
    
    Few-shot Steering (En.) & 0.8115 & \textbf{0.8666}$^{**}$ & 0.8596 & 0.8149 & 0.8629$^{**}$ & 0.8996$^{***}$ & 0.8775$^{***}$ & 0.8884$^{***}$ & 0.8601$^{***}$ \\ 
    \ \ +\textit{Language Steering} (Zh.) & \textbf{0.8469}$^{**}$ & 0.8652$^{**}$ & \textbf{0.8838}$^{***}$ & \textbf{0.8345}$^{**}$ & \textbf{0.8961}$^{***}$ & \textbf{0.9041}$^{***}$ & \textbf{0.8884}$^{***}$ & \textbf{0.8976}$^{***}$ & \textbf{0.8771}$^{***}$ \\

    \midrule
    \rowcolor{gray!10} \multicolumn{10}{c}{ \textit{Germany}} \\
    \midrule
    
    No Steering (En.) & 0.8175 & 0.8441 & 0.8541 & 0.7993 & 0.8468 & \textbf{0.8826} & 0.8280 & 0.8543 & 0.8408 \\ 
    \ \ +\textit{Language Steering} (De.) & \textbf{0.8445}$^{*}$ & \textbf{0.8494} & \textbf{0.8815}$^{**}$ & \textbf{0.8140} & \textbf{0.8765}$^{**}$ & 0.8801 & \textbf{0.8689}$^{**}$ & \textbf{0.8715} & \textbf{0.8608}$^{**}$ \\ 
    
    \hline
    
    Persona Steering (En.) & 0.8081 & \textbf{0.8510} & 0.8596 & 0.8193 & 0.8736$^{*}$ & 0.9129$^{***}$ & 0.8731$^{***}$ & 0.8879$^{***}$ & 0.8607$^{***}$ \\ 
    \ \ +\textit{Language Steering} (De.) & \textbf{0.8482}$^{*}$ & 0.8507 & \textbf{0.8904}$^{***}$ & \textbf{0.8250} & \textbf{0.8938}$^{***}$ & \textbf{0.9130}$^{***}$ & \textbf{0.8905}$^{***}$ & \textbf{0.9053}$^{***}$ & \textbf{0.8771}$^{***}$ \\ 
    
    \hline
    
    Few-shot Steering (En.) & 0.8345 & 0.8348 & 0.8619 & 0.8352$^{*}$ & 0.8670 & 0.8983 & 0.8709$^{***}$ & 0.8581 & 0.8576$^{***}$ \\ 
    \ \ +\textit{Language Steering} (De.) & \textbf{0.8661}$^{***}$ & \textbf{0.8683} & \textbf{0.8916}$^{***}$ & \textbf{0.8476}$^{**}$ & \textbf{0.9097}$^{***}$ & \textbf{0.9219}$^{***}$ & \textbf{0.9072}$^{***}$ & \textbf{0.9103}$^{***}$ & \textbf{0.8903}$^{***}$ \\

    \bottomrule
    \end{tabular}
    \caption{Culture representation scores for China and Germany under different steering methods for LLMs. 
    The content in ``()'' is to indicate the language being used, where ``En.'' denotes English, ``Zh.'' denotes Chinese, and ``De.'' denotes German. In all cases, the language of our inputs (task instruction, few-shot examples, and question) always keeps the same.
    Moreover, the significance is assessed using the $t$-test: * denotes $p$-value $<$ 0.05, ** denotes $p$-value $<$ 0.01, and *** denotes $p$-value $<$ 0.001.}
    \label{tab:steerability_results}
\end{table*}

\paragraph{Baselines}
Steerability refers to the ability of LLMs to adjust and align with the opinion of a target demographic group~\cite{meister-etal-2025-benchmarking}. 
Existing steering methods involve prepending additional context to the prompt describing the group we want the model to emulate. We consider the following steering methods as our baselines.

\begin{itemize}
    \item \textbf{Persona Steering} operates by inserting a concise persona description into the original task prompt, prompting the LLM to first ``locate'' itself as a member of a specific group and thereby explicitly emulate that group's opinion orientation.
    In this paper, we follow~\citet{santurkar2023whose} and~\cite{meister-etal-2025-benchmarking}, in which the LLM is instructed to pretend to be a member of the target country.
    \item \textbf{Few Shot Steering}~\cite{meister-etal-2025-benchmarking} provides the persona steering setting while also giving five in-context examples of the real group opinion distributions and asking the LLM to emulate the group's responses. This setting is representative when we already have the opinion distributions of the target group.
\end{itemize}

\paragraph{Method}
Language affects the way we think~\cite{Turner2000}, that is, speakers of different languages end up with different conceptual structures, and these differences can affect their worldview~\cite{plebe2015language}.
However, it is unclear how language affects the views of LLMs.
In this section, we validate the \textbf{language steering}, which enables an LLM to represent the opinions of speakers of language $l$ by changing the language of the prompt from English to non-English language $l$.
Specifically, for the task instructions, we translate them using GPT-4 and manually check the accuracy of the translated versions; for the in-context examples, we obtain the translations from the corresponding translated versions of the WVS questionnaires.
Unlike previous work~\cite{durmus2023towards} that only evaluates three languages, uses log probabilities of one LLM, and validates the language steering itself, we evaluate eight languages, use the verbalized distribution of seven LLMs, and validate the effect of combining the language steering with other steering methods.

Because WVS provides a multilingual translated version of the survey questionnaire for countries that use multiple languages (e.g., Singapore uses Chinese, English, and Malay; respondents may see the questionnaire in both English and Malay), the survey results from these countries may not reflect the opinions of single-language speakers.
Therefore, in order to quantitatively analyze the effect of language steering, we must avoid respondents using multilingual questionnaires. The languages and countries we choose to investigate must satisfy the following conditions: \textit{1)} the country must be surveyed by WVS using a single language; and \textit{2)} the language must be supported by the LLM.
After our selection, our experiments cover Spanish, Chinese, Japanese, Korean, German, Russian, Vietnamese, and Portuguese. As WVS covers survey data from 12 Spanish-speaking countries, we select three countries, Argentina, Chile, and Uruguay, to validate language steering. For other languages, we select one country for each language.

\paragraph{Effectiveness of Steering}

Table~\ref{tab:steerability_results} shows the alignment scores for China and Germany under different steering methods for LLMs.\footnote{The results of other languages are available in Appendix~\ref{sec:more_results_of_rq2}.} We show these two countries (or languages) due to their representation of non-Latin (Chinese) and Latin writing systems (German), respectively, forming a striking contrast in terms of language families and verifying the universality of the steering methods.
We can see that the baseline methods persona steering and few-shot steering show a certain steering ability on all countries compared to no steering.
In general, few-shot steering exhibits higher steering ability than persona steering, and our findings are consistent with existing studies~\cite{meister-etal-2025-benchmarking, studdiford2025evaluating}.
Moreover, ``few-shot + language steering'' achieves the highest alignment scores on both countries, which indicates the effectiveness of language in steering LLMs to emulate the opinions of language speakers.
Although persona steering can be inaccurate~\cite{meister-etal-2025-benchmarking}, ``persona + language steering'' also yields promising steering ability on China and Germany.
In addition, among the LLMs tested, DeepSeek-R1 exhibits the strongest response to the composite strategy, reflecting its greater ability to understand the countries we tested. 
Overall, language steering is effective in steering LLMs to emulate country-specific opinions.
This result implicates that while chasing higher benchmark scores may suggest using a language other than the target one~\cite{kaneko-etal-2025-balanced}, for opinion-oriented tasks, sticking to the target language can yield better alignment. Therefore, whenever we consider deploying other languages, we must pay careful attention to value alignment and similar factors.

\section{Temporal Opinion Alignment (RQ3)}
Human cognition has a social historical character~\cite{berger2016social}, which means that human cognition is not static; it is constantly being reshaped as historical conditions evolve~\cite{vygotsky2012thought, mannheim2013ideology}. Therefore, it is important to know whether LLMs reflect contemporary human opinions.
In this study, we suppose that \textit{it is desirable for LLMs to reflect contemporary human opinions}. Next, we will show the representativeness of LLMs with human opinion distributions in the temporal dimension.

\paragraph{Method}
As shown in the results of Figure~\ref{fig:rq1_whole}(c), the LLM is under-aligned with most countries' opinions. We therefore consider which countries are well aligned by the LLM for comparison. Formally, we use the following equation to filter the countries for comparison:
\begin{equation}
    \mathcal{C}^{*} = \{\, c \in \mathcal{C} \mid \left| \mathcal{A}_{\mathcal{M}}^{(c)} - \mathcal{A}_{avg}^{(c)} \right| < \tau \,\}.
\label{eq:filter}
\end{equation}
where $\mathcal{A}_{\mathcal{M}}^{(c)}$ denotes the alignment score between the LLM's opinion distribution and that of country $c$. Likewise, $\mathcal{A}_{avg}^{(c)}$ is the alignment score between the average human opinion distribution and that of country $c$. The margin $\tau$ is the threshold, which we set to 0.02 to ensure that we can filter out at least five countries that are well aligned by LLMs.
We refer readers to Appendix~\ref{sec:more_results_of_rq3} for more details on filtering and filtered countries.

\paragraph{Results} 

\begin{figure}[t]
\centering
\includegraphics[width=\columnwidth]{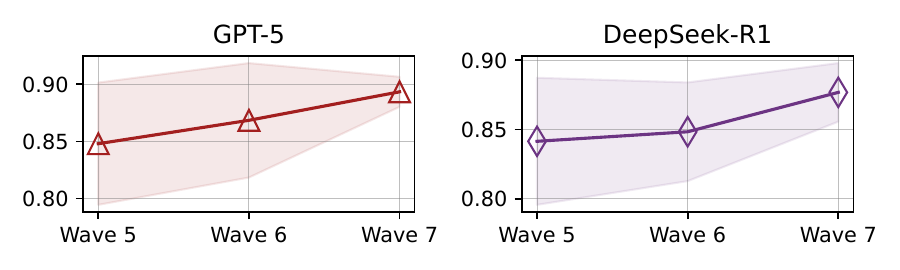}
\caption{The trend of the average alignment scores of GPT-5 and DeepSeek-R1 with countries filtered using Eq.~(\ref{eq:filter}) across waves. The shaded area indicates the standard deviation of the alignment scores at the current wave.}
\label{fig:historical_period}
\end{figure}

Figure~\ref{fig:historical_period} illustrates the trend of the average alignment scores of GPT-5 and DeepSeek-R1 across waves.
We can see that LLMs most align with the latest wave of human opinions. 
One reason may be due to the fact that most of the human feedback in the fine-tuning of LLMs comes from contemporary annotators, whose ethical and social orientations are more in line with recent public opinions. The models are rewarded for emulating these behaviors.
Another reason may be due to the development of mobile internet and social media, which provide more accessible data than in the past.
In addition, the standard deviation reflects an LLM's coverage degree of the represented opinions. A lower standard deviation indicates a better coverage degree of the opinions. Therefore, LLMs achieve better coverage of human opinions in the latest wave and poorer coverage of human opinions in earlier waves. This further indicates that the alignment of LLMs is influenced by contemporary annotators' human feedback. In contrast, it implies that there may be challenges for LLMs to align to earlier human opinions. The reason is that it is more difficult to access earlier customized human feedback data.

\section{Discussion}

\subsection{Alignment of Human Opinions across Countries}
\begin{figure}[t]
\centering
\includegraphics[width=\columnwidth]{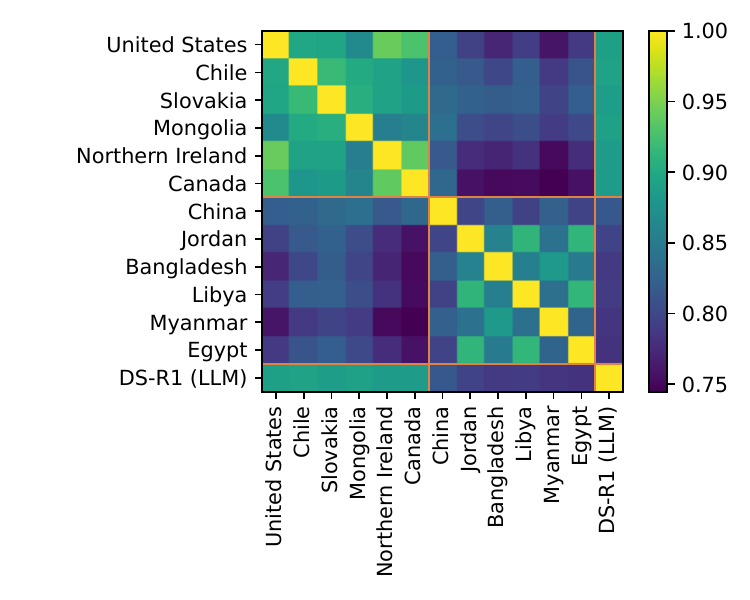}
\caption{Alignment scores among different countries. For comparison, we also show the alignment scores between the countries and DeepSeek-R1 (DS-R1).}
\label{fig:human_performance}
\end{figure}

To understand the alignment between different countries' opinions, we show the alignment scores between countries and countries in Figure~\ref{fig:human_performance}. 
Similar to Figure~\ref{fig:rq1_whole}(b), we plot the first and last 6 countries ranked by their alignment scores with DeepSeek-R1's opinion distributions.
For comparison, we also show the alignment scores between the countries and DeepSeek-R1 in the results.
More results can be found in Appendix~\ref{sec:alignment_of_human_opinions_across_countries}.
The alignment scores of opinion distributions among the first 6 ranking countries are generally high, indicating significant similarity in their views.
In contrast, the alignment scores among the last 6 ranking countries show relative differences. For example, Jordan, Libya, and Egypt have high alignment scores with each other, while Bangladesh, China, and Myanmar have low alignment scores with other countries.
Meanwhile, the alignment scores between the first and last 6 ranking countries are generally low. 
Furthermore, by calculating the alignment scores between DeepSeek-R1 and each country, we find that despite some heterogeneity in alignment scores among the first 6 ranking countries, DeepSeek-R1's alignment scores with these countries are surprisingly consistent. This pattern suggests that the model captures an ``average'' cross-country view rather than aligning specifically to individual countries, revealing a lack of country-specific sensitivity in the alignment of opinion distributions.

\subsection{Internal Consistency of LLMs} 
In order to investigate whether LLMs hold self-contradictory opinions when answering questions, we conduct a closer look at the internal consistency of LLMs.
Concretely, we manually select four questions about gender fairness (agree or disagree that men are superior to women), four questions about atheism (believe or disbelieve in theism), and two questions about democracy (important or not important that democracy).\footnote{A complete list is provided in Appendix~\ref{sec:internal_consistency_of_llms}.}
Questions with the same topic reflect the same logic of opinion judgment.
In our experiments, we consider the options ``Strong agree'' and ``Agree,'' which both express positive attitudes, to be consistent opinions.
For options related to degree (e.g., from ``1. Not at all important'' to ``10. Absolutely important''), we group options 1 to 5 and options 6 to 10 as consistent opinion groups, respectively. 

We use the internal consistency rate, which is the ratio of answers within the same opinion group, as the evaluation measure.
If the answers to a set of questions are [1, 1, 2, 1], the internal consistency rate is 75\%.
The experimental results show that on the topic of gender fairness, all the LLMs remain at a 100\% internal consistency rate.
However, GPT-3.5-Turbo shows only 50\% internal consistency on the topics of atheism and democracy, in contrast to other LLMs, which maintain full consistency.
This suggests that, relative to other LLMs, GPT-3.5-Turbo not only poorly aligns with human opinions (\S\ref{sec:global_opinion_representation}), but also lacks internal consistency.

\subsection{Sensitivity of LLMs}
\begin{table}[t]
    \centering
    % \small
    \begin{tabular}{lccc}
        \toprule
            \textbf{Models} &   \textbf{Order}  & \textbf{\#few-shot} & \textbf{Prob.} \\
            \midrule
              Aya23         & 0.9660 & 0.9893 & 0.9815 \\
              Llama3        & 0.9927 & 0.9964 & 0.9933 \\
              Qwen2.5       & 0.9928 & 0.9928 & 0.9973 \\
              GPT-3.5       & 0.9852 & 0.9810 & 0.9736 \\
              GPT-4         & 0.9925 & 0.9783 & 0.9902 \\
              GPT-5         & 0.9973 & 0.9968 & 0.9980 \\
              DeepSeek-V3   & 0.9895 & 0.9805 & 0.9881 \\
              DeepSeek-R1   & 0.9885 & 0.9861 & 0.9924 \\
        \bottomrule
    \end{tabular}
    \caption{Pearson's $r$ between the default and sensitivity test results. Here, the default result means the alignment scores between LLMs' opinion distributions and those of different countries under default prompts. The sensitivity test result means the alignment scores between LLMs' opinion distributions and those of different countries under sensitivity test prompts. \textbf{Order} means shuffle the input question's options, \textbf{\#few-shot} means limit the number of the few-shot examples from five to three, and \textbf{Prob.} means change the probability distribution of the few-shot examples. \textit{Note}: The $p$-values for all results are less than 0.001.}
    \label{tab:sensitivity_of_llms}
\end{table}

We acknowledge that current LLMs are very sensitive to the prompt format~\cite{zhuo2024prosa}.
Inspired by~\citet{santurkar2023whose}, we test the prompt sensitivity of the LLMs used in our experiments against the following factors:
\begin{itemize}
    \item The order in which the question options are presented to the model;
    \item The few-shot examples (the number of the few-shot examples and the probability distribution of the few-shot examples).
\end{itemize}
We employ the Pearson's $r$ of the alignment scores of different sensitivity test prompts for evaluating prompt sensitivity. As shown in Table~\ref{tab:sensitivity_of_llms}, despite changing prompts, the results output by LLMs still maintain a significantly high Pearson's $r$, and the $p$-value is less than 0.001. Moreover, Appendix~\ref{sec:sensitivity_of_llms}, \ref{sec:g4}, \ref{sec:g5} and \ref{sec:g6} provide more results and discussions.

\section{Related Work}
\paragraph{Human Opinion and Culture.}
Culture can be defined as the pattern of thinking, feeling, and reacting, distinguishing human groups~\cite{wallace1961culture,shweder2007cultural,cao-etal-2023-assessing}. 
Human opinions are the beliefs, preferences, and judgments held by individuals, and are deeply influenced by cultural backgrounds~\cite{peterson2003culture, markus2014culture}.
Culture provides people with a shared framework through which they interpret experiences, internalize norms, and express their opinions~\cite{geertz2017interpretation}. It is challenging to align the opinions of human societies with different cultural backgrounds~\cite{hall1976beyond}.

\paragraph{Culture in LLMs.}
Previous studies have shown that LLMs exhibit cultural challenges similar to those found in human society~\cite{li2024culturellm}.
\citet{liu-etal-2024-multilingual} pointed out that there is a cultural gap when using LLMs, especially when dealing with content translated from other languages.
After that, \citet{masoud-etal-2025-cultural} demonstrated that GPT-4 exhibits remarkable capabilities in adapting to these cultural gaps, particularly in the Chinese context.
Furthermore, \citet{wang-etal-2024-countries} showed that LLMs have cultural dominance issues because they are mainly trained using English data. However, recent study~\cite{meister-etal-2025-benchmarking} has challenged the existing use of LLMs to emulate opinion distributions, proving that LLMs are more accurate when describing opinion distributions. Therefore, one technical difference from existing studies is that we deployed this method of describing opinion distribution (known as \textit{verbalized distribution}).

\paragraph{Opinion of LLMs.}
LLMs often respond to subjective questions in a way that implicitly favors certain human groups. Early study~\cite{santurkar2023whose} quantified the differences between LLMs' predicted distributions of response options and human distributions based on surveys. \citet{meister-etal-2025-benchmarking} expanded the assessment scope to multiple demographic dimensions and compared different distribution expressions (model log-probability, sequence of tokens, and verbalized distribution). Building on this research direction, this study expands the scope from a few countries to a truly worldwide country sample and explicitly introduces the temporal dimension.

\paragraph{Steering Methods.}
To enable LLMs to emulate specific groups, previous studies~\cite{meister-etal-2025-benchmarking} have explored persona and few-shot steering. However, persona steering is not always accurate, leading to undesirable effects such as stereotypes and exacerbated polarization~\cite{perez-etal-2023-discovering,cheng-etal-2023-marked, wang2025large}.
We further expanded on these two methods and proposed language steering: changing the language of the prompt can significantly improve alignment with the opinion distribution of the corresponding country, indicating that language itself, as an alignment cue, can drive opinion alignment.

\section{Conclusion}

In this paper, we propose a framework to investigate the global human opinion alignment of LLMs using the WVS questionnaires and data.
We manually filter the 259 questions in the WVS questionnaires across multiple language versions and historical periods to build our dataset.
Through experimenting with LLMs on our data, our main findings are as follows:
\textit{1)} LLMs only appropriately align with a few countries, while under-aligning with most countries.
\textit{2)} Language can significantly steer LLMs to emulate country-specific opinions.
\textit{3)} LLMs most align with the contemporary human opinions.
Our study also discusses various aspects of LLMs, such as internal consistency and sensitivity of LLMs.
This paper will provide insights for studies related to the value alignment of LLMs.

\section*{Acknowledgments}
We thank the reviewers for their valuable feedback.
This work was supported by JST BOOST, Grant Number JPMJBS2407.

\bibliography{aaai2026}

% \newpage
% \input{checklist/ReproducibilityChecklist}

\newpage
\appendix
\section{More Details of The World Values Survey}
\label{sec:more_detals_of_wvs}
Collecting data on human opinions across global, language, and temporal dimensions is costly.
The WVS is an international academic project studying human values, beliefs, norms, self-descriptions, and other elements of national culture, as well as political attitudes and opinions, across the globe. The scholars who lead the project are mainly interested in the relationships between economic development, cultural change, and political life on all continents. The WVS data have also been widely used by cross-cultural psychologists, anthropologists, and other social scientists~\cite{minkov2012world}.
The project started in 1981 and has been operating in more than 120 world countries and regions. The main research instrument of the project is a representative comparative social survey which is conducted globally every 5 years~\cite{wvs_round7_v6}.
The latest wave of the survey employs a standardized questionnaire containing approximately 259 subjective questions. Its broad regional and topical coverage, as well as its free public access to survey data and research results, makes the WVS one of the most authoritative and widely used multi-country surveys in the field of social sciences.

\section{Experiment Settings}
\label{sec:experiment_settings}
\subsection{Hyperparameter}
The hyperparameter settings for the LLMs (if available) are shown in Table~\ref{tab:hyper_parameters}.
\begin{table}[H]
    \centering
    \begin{tabular}{ll}
    \textbf{Hyperparameter} & \textbf{Assignment} \\
    \midrule
    Top-$p$ & 1.0  \\
    Temperature & 0.0 \\
    Max new tokens & 256 \\
    Frequency penalty & 0.0 \\
    Presence penalty & 0.0 \\
    \end{tabular}
    \caption{Names and descriptions of columns in \texttt{jsonl} file.}
    \label{tab:hyper_parameters}
\end{table}
For GPT-5, we set the reasoning configuration to \texttt{reasoning = \{effort = ``low''\}} and the text generation configuration to \texttt{text = \{verbosity = ``low''\}} to obtain concise responses with reduced reasoning depth.

\subsection{Question Filter Rules}
The wave 7 questionnaire contains 259 questions, but not all of them are suitable for LLMs to answer. If the questions are not selected appropriately, it may cause bias in the evaluation of LLMs.  
Therefore, we first filter the questions.
We filter out the following types of questions from the WVS questionnaire:

\paragraph{Not multiple choice.} 
For example, questions Q7 to Q17 each contain two options (``Mentioned'' and ``Not mentioned''), and the questionnaire requires respondents to choose up to 5 mentioned options from Q7 to Q17. 

\paragraph{Require life experience to answer.}
Some questions require life experience to answer. Therefore, they are not suitable for LLMs to answer. For example, questions Q51 to Q55 begin with ``In the last 12 months.''

\paragraph{Need to edit slot.} Some questions need to edit a slot according to the respondent's country. For example, Q64 needs to modify the slot \texttt{[churches]} according to the respondent's country.
In the ``No steering'' scenario, we do not want to leak any country information to LLMs.

\paragraph{Objective question.}
Some questions are objective questions. For example, question Q91, ``\textit{Five countries have permanent seats on the Security Council of the United Nations. Which one of the following is not a member? A) France, B) China, C) India}.''

\paragraph{Require a nationality.}
Some questions require the respondents to have a nationality, so they are not suitable for LLMs to answer. For example, question Q120, ``\textit{How high is the risk in this country to be held accountable for giving or receiving a bribe, gift or favor in return for public service? To indicate your opinion, use a 10-point scale where `1' means `no risk at all' and `10' means `very high risk'}.''

\paragraph{}
We put all the filtered 144 questions in the \texttt{WV7\_English.jsonl} file under the \texttt{dataset/questions} folder.
The naming convention for saving files with questions in other languages is to replace \texttt{English} with the other language (e.g., \texttt{German}).
Table~\ref{tab:column_explain} is the explanation of each column in the \texttt{jsonl} file.
\begin{table}
    \centering
    \begin{tabular}{lp{4cm}}
    \textbf{Names} & \textbf{Descriptions} \\
    \midrule
    \texttt{id} & The question id  \\
    \texttt{question} & The question description \\
    \texttt{choice\_keys} & The option number \\
    \texttt{choices} & The option content \\
    \texttt{answer} & The combined option number and option content \\
    \end{tabular}
    \caption{Names and descriptions of columns in \texttt{jsonl} file.}
    \label{tab:column_explain}
\end{table}

\begin{table}[t]
    \centering
    \small
    \begin{tabular}{ll}
            \textbf{Country} &   \textbf{Example IDs}  \\
            \midrule
            Default (English version) & Q60, Q70, Q90, Q110, Q130 \\
            Argentina & Q40, Q60, Q80, Q150, Q160 \\
            Brazil & Q40, Q60, Q80, Q150, Q160 \\
            Chile & Q40, Q60, Q80, Q150, Q160 \\
            China & Q60, Q70, Q110, Q150, Q160  \\
            Germany & Q40, Q80, Q150, Q160, Q170 \\
            Japan & Q60, Q70, Q90, Q110, Q130 \\
            Korea & Q40, Q80, Q150, Q160, Q170 \\
            Russia & Q40, Q60, Q80, Q150, Q160 \\
            Uruguay & Q40, Q60, Q80, Q150, Q160 \\
            Vietnam & Q40, Q60, Q80, Q150, Q160 \\
    \end{tabular}
    \caption{The question ids of the few-shot examples we selected for each country.}
    \label{tab:few_shot_ids}
\end{table}

\subsection{Full Prompts}

We follow the settings of~\citet{meister-etal-2025-benchmarking} by providing few-shot examples to control the LLMs' output format. Our prompts follow the following format:

\begin{tcolorbox}[
  colback=white,           
  colframe=black,      
  boxrule=1pt,
  arc=0pt,
  left=8pt,
  right=8pt,
  top=6pt,
  bottom=6pt,
  enhanced,
  breakable,
  fontupper=\rmfamily
]
% \small
\textbf{Task instruction:} \\ I will provide a distribution over answer choices on a series of questions to help you understand the formatting of this task. After the 5 examples, please express the distribution of answers for the question asked. Do not follow the distribution expressed in the examples as they are only to give you an example of formatting, express the distribution based on your understanding of how humans may respond. Please only respond in the exact format of a dictionary mapping answer choice number to probability, no extra spaces, characters, quotes or text. Please only produce 1 sentence in this format. Answers outside of this format will not be accepted. First I will provide a few examples of this task just to give you a sense of the answer formatting, then you will perform the task yourself.\\
\\
\textbf{Few-shot examples:} \\
Question: How much do you trust people you know personally? \\
`1'. Trust completely \\
`2'. Trust somewhat \\
`3'. Do not trust very much \\
`4'. Do not trust at all \\
Answer: \{`1': `31.01\%', `2': `3.21\%', `3': `30.31\%', `4': `35.47\%'\}\\
\\
\textit{(The other four examples have been omitted here)} \\
\\
\textbf{Input question:} \\ 
Question: How important is family in your life? \\
`1'. Very important \\ 
`2'. Rather important \\ 
`3'. Not very important \\ 
`4'. Not at all important\\ 
Answer: \\
\textit{(We can get a distribution in JSON format)}

\end{tcolorbox}

The task instruction part will slightly differ depending on the steering method. For example, in the case of persona steering, the task instruction will require the LLMs to emulate people from a specific country (e.g., Germany).
Note that in the case of language steering, the task instruction, few-shot examples, and input question are all replaced with questions in the target language or translated into the target language.

When selecting few-shot examples, there may be missing opinion distributions for some questions depending on the country. Table~\ref{tab:few_shot_ids} lists the question ids of the few-shot examples we selected for each country.

In addition, all few-shot examples are available under the \texttt{input\textbackslash express\_distribution} folder. Table~\ref{tab:few_shot_file} lists each file and its corresponding explanation.
\begin{table*}[h]
    \centering
    \begin{tabular}{ll}
            \textbf{File Name} &   \textbf{Steering Method}  \\
            \midrule
            \texttt{lang-En\_dist-random.txt} & no steering, persona steering \\
            \texttt{lang-En\_dist-\{country\}.txt} & few-shot steering \\
            \texttt{lang-\{language\}\_dist-random.txt} & \makecell[l]{language steering\\ persona + language steering} \\
            \texttt{lang-\{language\}\_dist-\{country\}.txt} & few-shot + language steering \\
    \end{tabular}
    \caption{The few-shot examples under the \texttt{inputs/few\_shot} folder. Here, \texttt{\{language\}} represents the language of the prompt, which can be De (German), En (English), Es (Spanish), Ja (Japanese), Ko (Korean), Pt (Portuguese), Ru (Russian), Vi (Vietnamese), and Zh (Chinese). The \texttt{random} means that the distribution of examples is random, and \texttt{\{country\}} means that the distribution of examples is the opinion distribution of that country.}
    \label{tab:few_shot_file}
\end{table*}

\subsection{More Details of Distribution Expression Methods}

\paragraph{Model log probability.} A common method for estimating the choice distribution is to consider the model's log probabilities of the next token for each option token (e.g., `1', `2', etc.) as a class distribution and sample from it. This method is a standard practice~\cite{santurkar2023whose}. However, previous studies have shown that the probabilities derived from LLMs are usually much sharper than real human opinion distribution, i.e., most of the probabilities are concentrated in a few answers~\cite{durmus2024towards}.

\paragraph{Sequence of tokens.}
Another method is to directly let the model ``act as a sampler.'' Specifically, we instruct the LLM to output 30 samples from its internal distribution at once (e.g., \texttt{1124243412434213}). This method is more convenient when generating samples from the opinion distribution for emulation. However, since we use a limited-length sequence to approximate the underlying distribution, the estimation accuracy may be subject to errors due to limits on sample size~\cite{meister-etal-2025-benchmarking}.

\paragraph{Verbalized distribution.} This method directly allows the LLM to verbalize the distribution in JSON format in their output (e.g., \texttt{\{1: 31\%, 2: 4\%, 3: 30\%, 4: 35\%\}}), without any estimation steps or post-processing~\cite{meister-etal-2025-benchmarking}.
\citet{meister-etal-2025-benchmarking}'s analysis reveals that ``verbalized distribution'' outperforms the other ``model log probability'' and ``sequence of tokens''.
Therefore, we use ``verbalized distribution'' to represent the opinion distributions in our study.

\section{Complete Country List}
\label{sec:complete_country_list}
Wave 7 of the WVS covers 66 countries or regions. Here is the complete list, with the languages supported in parentheses. 
The contents in ``()'' are the language(s) in which the questionnaire is available for each country.

Andorra (Catalan, English, French, Spanish), Argentina (Spanish), Armenia (Armenian), Australia (English),
Bangladesh (Bangla), Bolivia (Spanish), Brazil (Portuguese), Canada (English, French), Colombia (Spanish),
Cyprus (Greek, Turkish), Czechia (Czech), Chile (Spanish), China (Chinese), Ecuador (Spanish), Egypt (Arabic),
Ethiopia (Afan Oromo, Amharic, Tigrinya), Germany (German), Greece (Greek), Guatemala (Spanish), Hong Kong SAR (Chinese, English),
India (Bengali, English, Hindi, Marathi, Punjabi, Telugu), Indonesia (Indonesian), Iran (Farsi), Iraq (Arabic),
Japan (Japanese), Jordan (Arabic), Kazakhstan (Kazakh, Russian), Kenya (Swahili), Kyrgyzstan (Kyrgyz, Russian),
Lebanon (Arabic), Libya (Arabic), Macau SAR (Chinese), Malaysia (Chinese, Malay), Maldives (Dhivehi), Mexico (Spanish),
Mongolia (Mongolian), Morocco (Arabic), Myanmar (Burmese), Netherlands (Dutch), New Zealand (English), Nicaragua (Spanish),
Nigeria (Hausa, Igbo, Yoruba), Pakistan (Urdu), Peru (Spanish), Philippines (Bicol, Cebuano, Filipino, Hiligaynon, Iluko, Tausug, Waray),
Puerto Rico (Spanish), Romania (Romanian), Russia (Russian), Serbia (Serbian), Singapore (Chinese, English, Malay),
Slovakia (Slovak), South Korea (Korean), Taiwan (Chinese), Tajikistan (Russian, Tajik), Thailand (Thai), Tunisia (Arabic),
Turkey (Turkish), Ukraine (Russian, Ukrainian), United Kingdom Great Britain (English), United Kingdom Northern Ireland (English),
Uruguay (Spanish), United States (English), Uzbekistan (Uzbek), Venezuela (Spanish), Vietnam (Vietnamese), and Zimbabwe (Ndebele, Shona).

\section{More Results of RQ1}
\label{sec:more_results_of_rq1}

\subsection{Country Level}
\label{sec:more_results_for_country_level}
We provide the alignment scores for all seven LLMs with different countries; the checklist is shown in Table~\ref{tab:compare_countries}.

\begin{figure}[t]
\centering
\includegraphics[width=0.85\columnwidth]{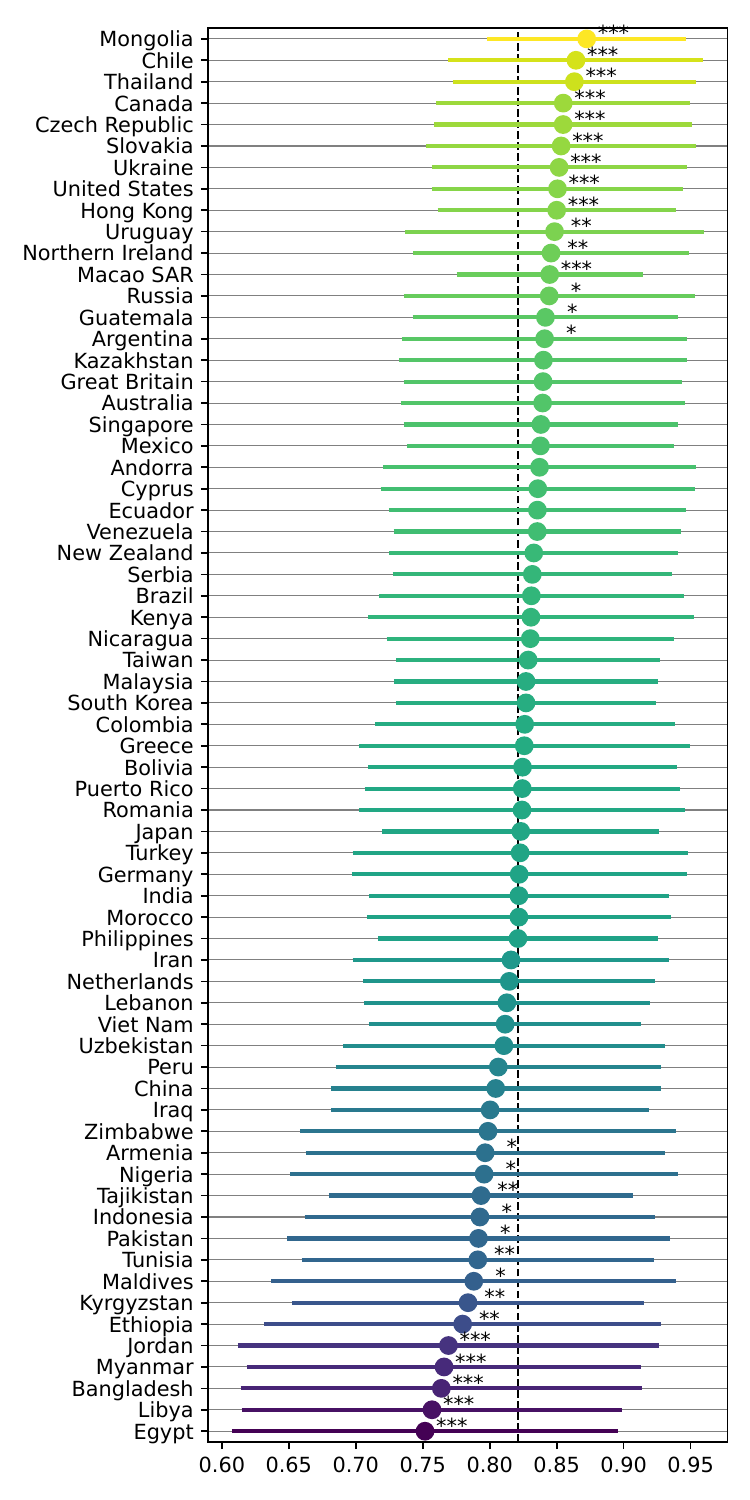}
\caption{The alignment scores between Aya23 and different countries.}
\label{fig:compare_countries_aya23}
\end{figure}

\begin{figure}[t]
\centering
\includegraphics[width=0.85\columnwidth]{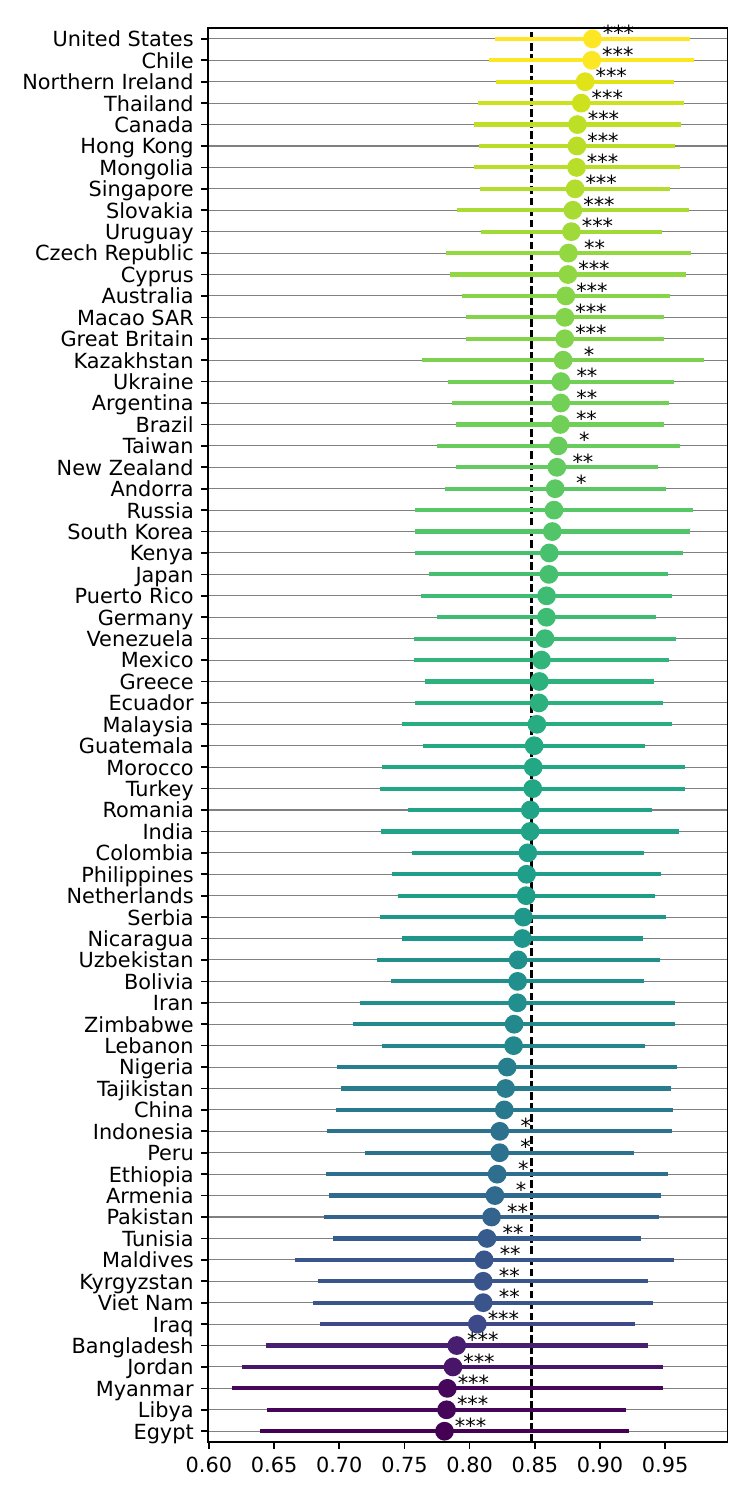}
\caption{The alignment scores between Qwen2.5 and different countries.}
\label{fig:compare_countries_qwen25}
\end{figure}

\begin{figure}[t]
\centering
\includegraphics[width=0.85\columnwidth]{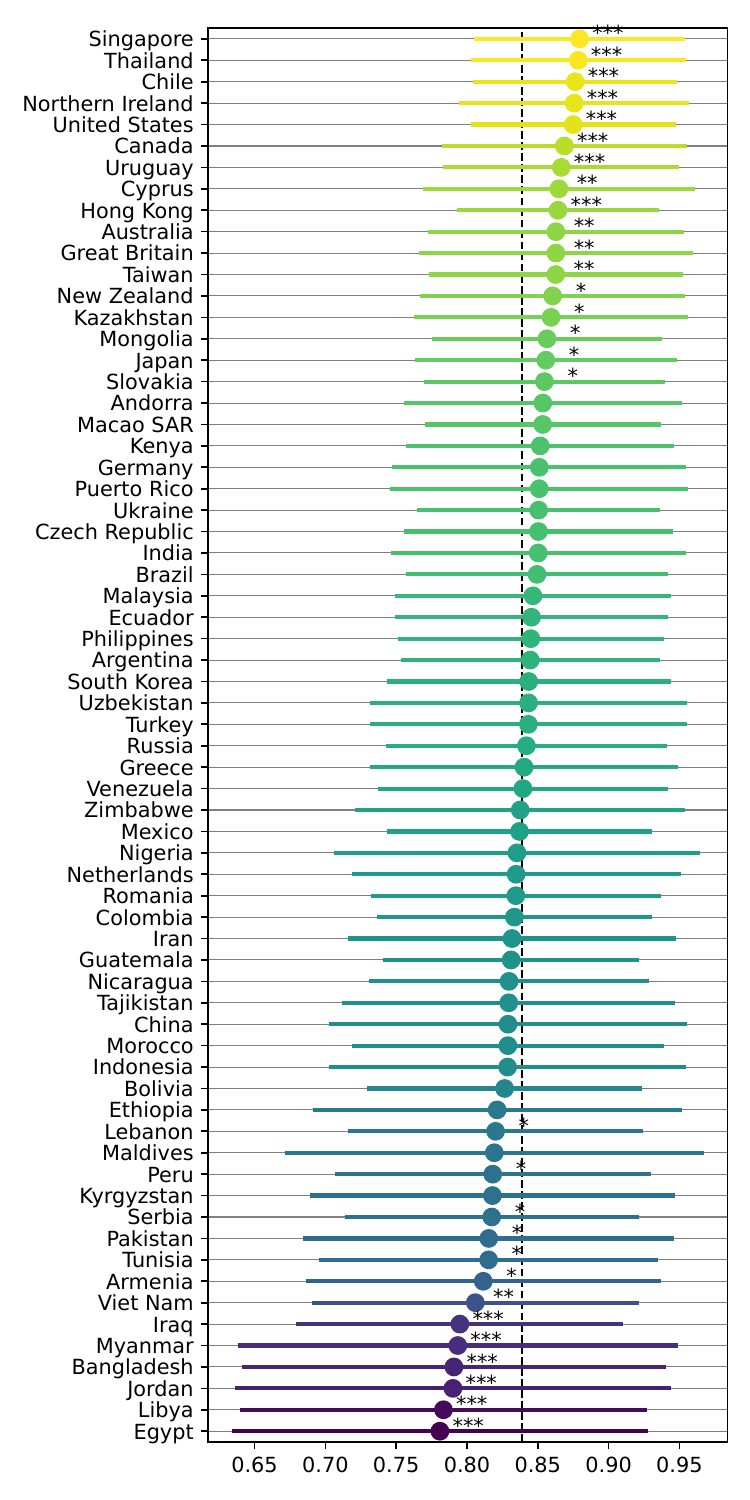}
\caption{The alignment scores between Llama3 and different countries.}
\label{fig:compare_countries_llama3}
\end{figure}

\begin{figure}[t]
\centering
\includegraphics[width=0.85\columnwidth]{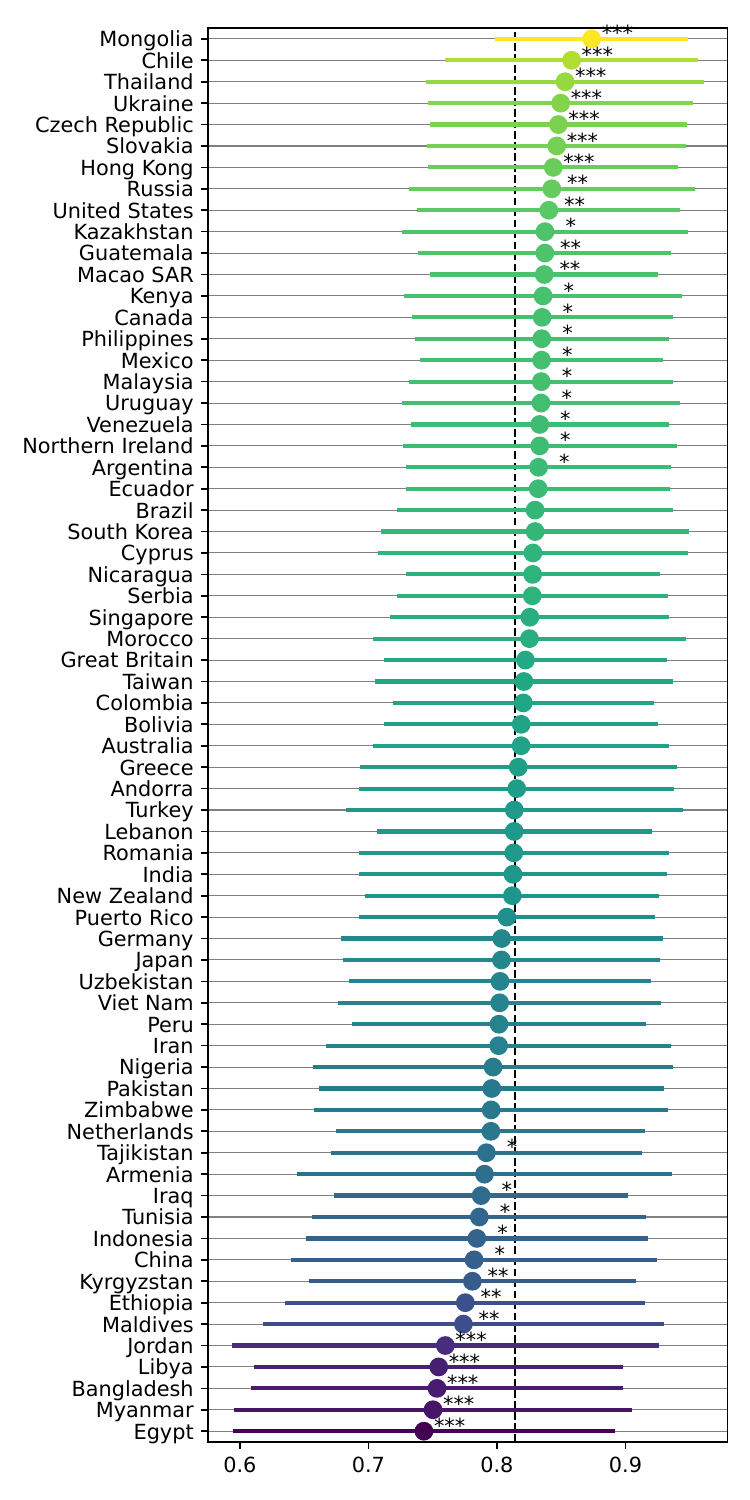}
\caption{The alignment scores between GPT-3.5 and different countries.}
\label{fig:compare_countries_gpt35}
\end{figure}

\begin{figure}[t]
\centering
\includegraphics[width=0.85\columnwidth]{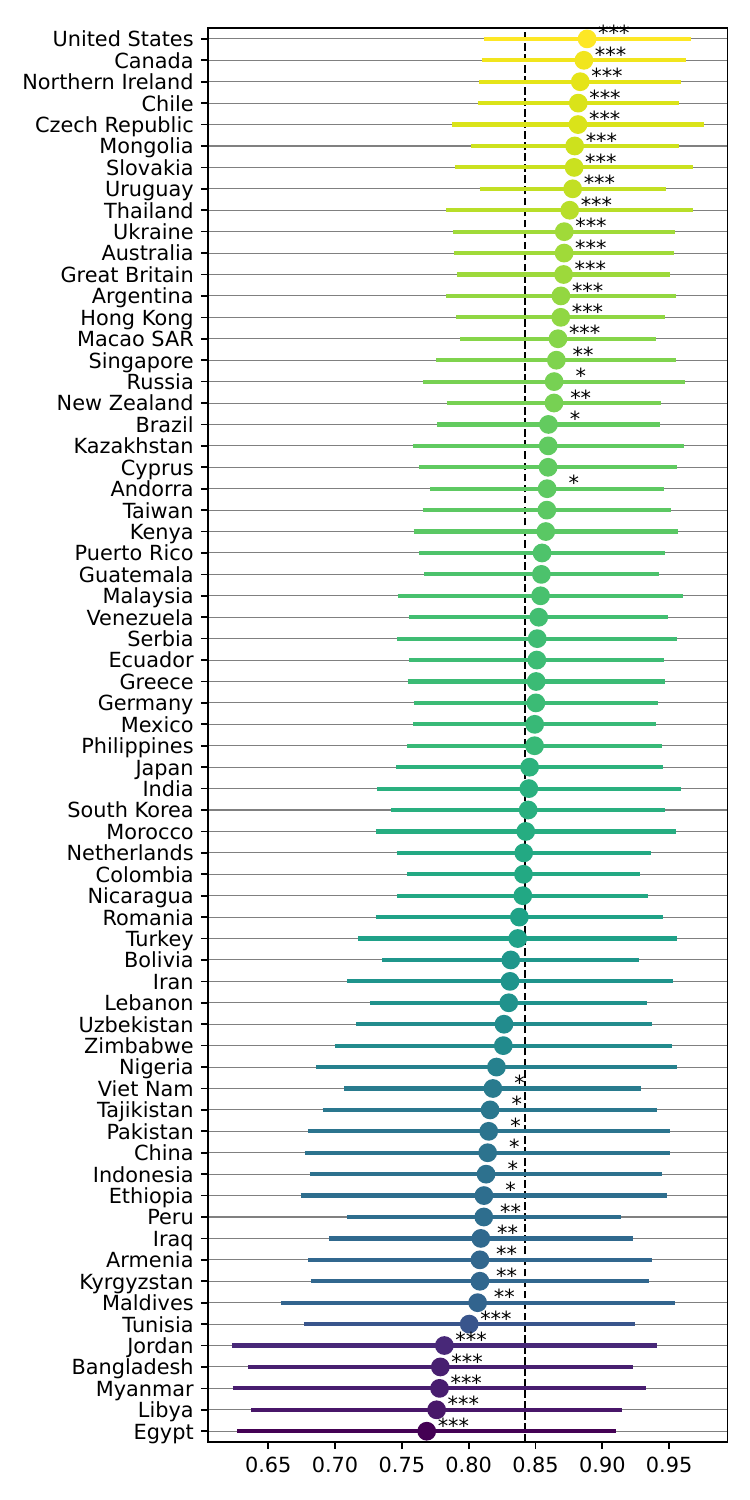}
\caption{The alignment scores between GPT-4 and different countries.}
\label{fig:compare_countries_gpt4}
\end{figure}

\begin{figure}[t]
\centering
\includegraphics[width=0.85\columnwidth]{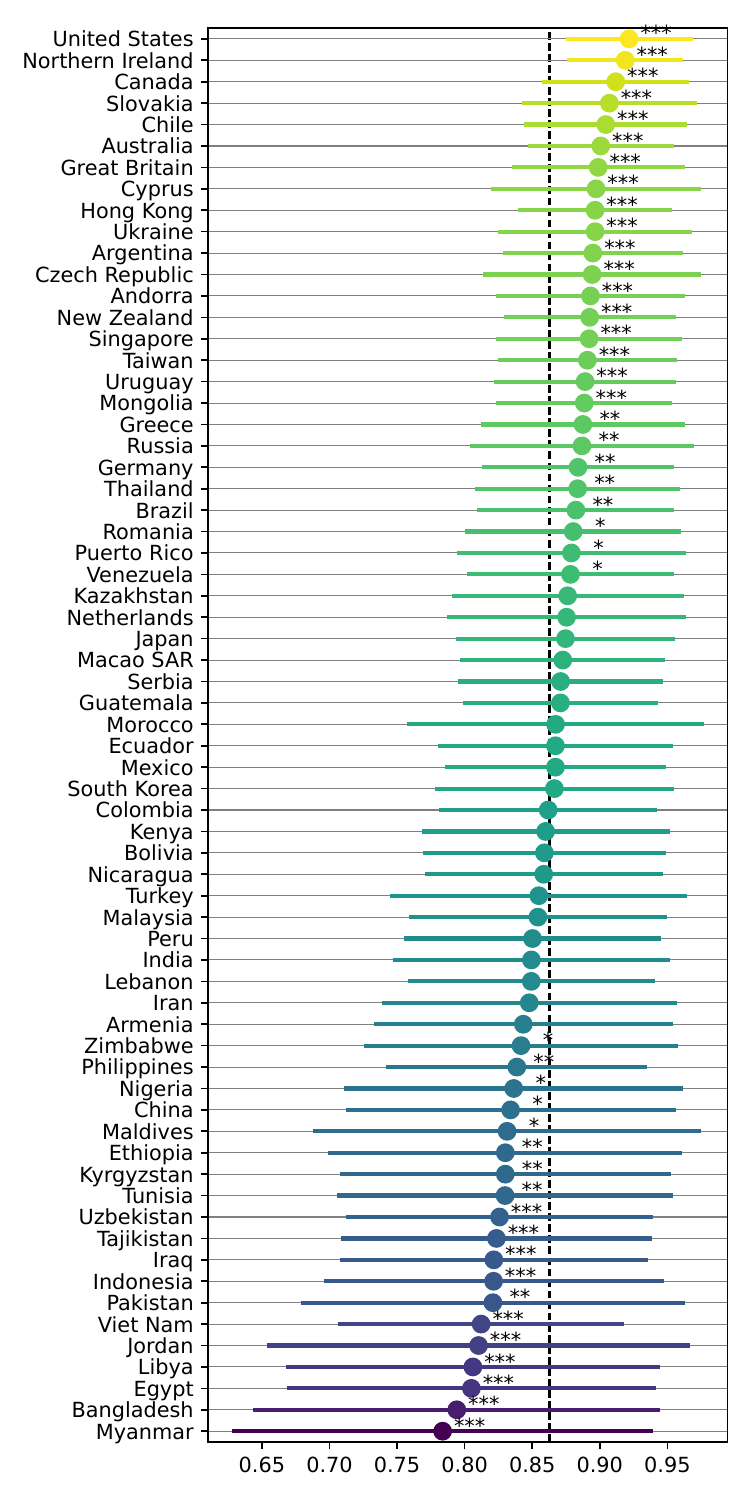}
\caption{The alignment scores between GPT-5 and different countries.}
\label{fig:compare_countries_gpt5}
\end{figure}

\begin{figure}[t]
\centering
\includegraphics[width=0.85\columnwidth]{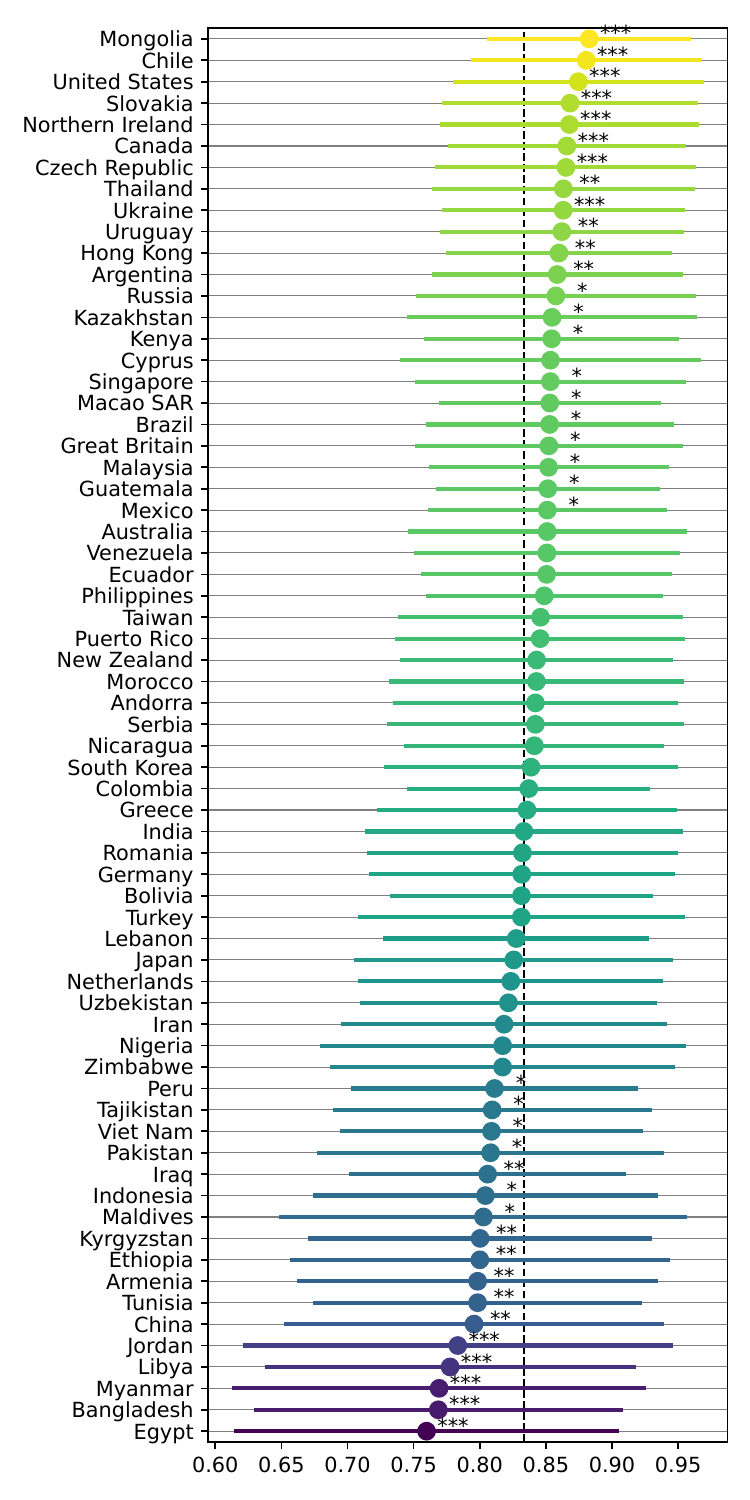}
\caption{The alignment scores between DeepSeek-V3 and different countries.}
\label{fig:compare_countries_dsv3}
\end{figure}

\begin{figure}[t]
\centering
\includegraphics[width=0.85\columnwidth]{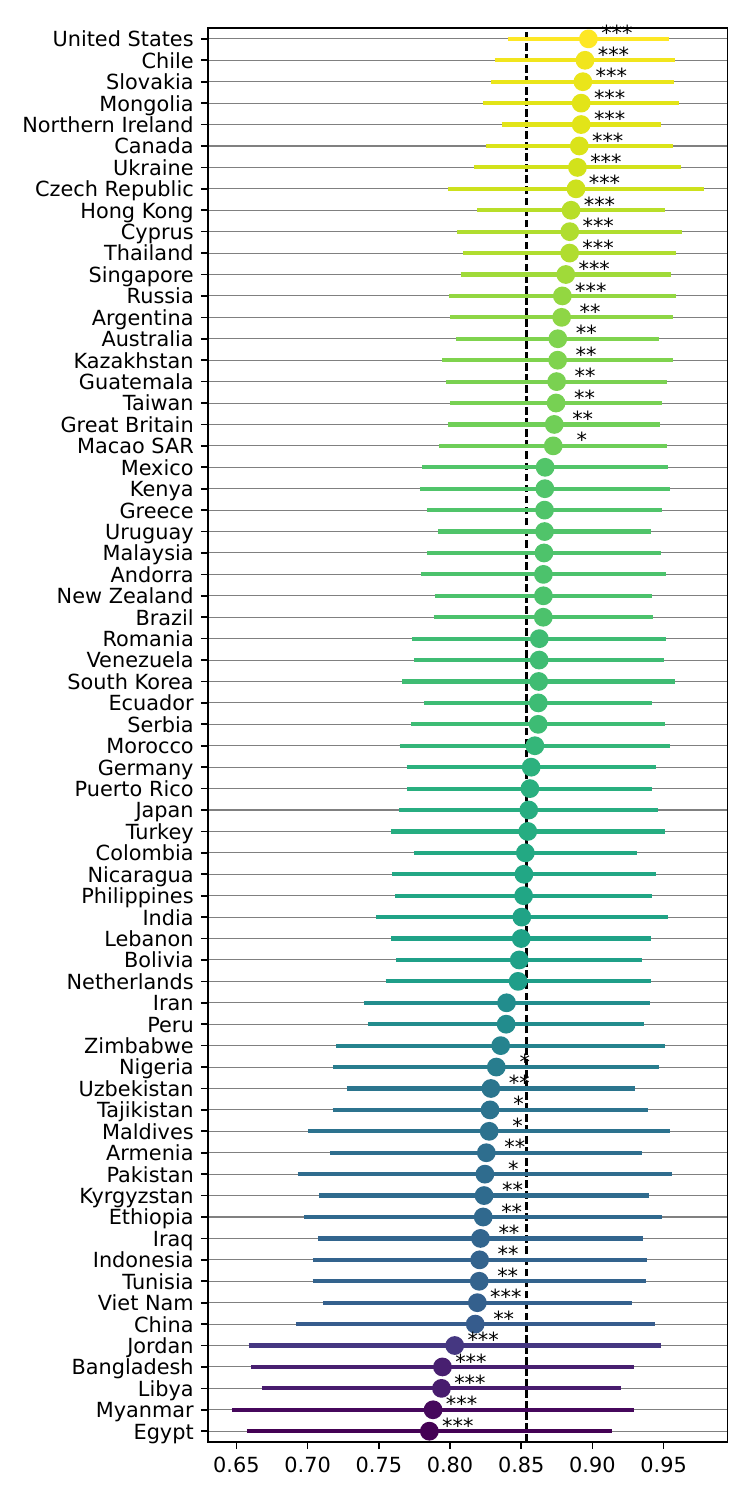}
\caption{The alignment scores between DeepSeek-R1 and different countries.}
\label{fig:compare_countries_dsr1}
\end{figure}

\begin{table}[h]
    \centering
    \begin{tabular}{ll}
            \textbf{Models} &   \textbf{Figures}  \\
            \midrule
            Aya23 & Figure~\ref{fig:compare_countries_aya23} \\
            Qwen2.5 & Figure~\ref{fig:compare_countries_qwen25} \\
            Llama3 & Figure~\ref{fig:compare_countries_llama3} \\
            GPT-3.5 & Figure~\ref{fig:compare_countries_gpt35} \\
            GPT-4 & Figure~\ref{fig:compare_countries_gpt4} \\
            GPT-5 & Figure~\ref{fig:compare_countries_gpt5} \\
            DeepSeek-V3 & Figure~\ref{fig:compare_countries_dsv3} \\
            DeepSeek-R1 & Figure~\ref{fig:compare_countries_dsr1} \\
    \end{tabular}
    \caption{Checklist of alignment scores between LLMs and different countries.}
    \label{tab:compare_countries}
\end{table}

By comparing Figures~\ref{fig:compare_countries_aya23} to~\ref{fig:compare_countries_dsr1}, we find that these models have high alignment scores with the United States, Chile, Thailand, Canada, Mongolia, and Slovakia. However, they have low alignment scores with Bangladesh, Myanmar, Libya, and Egypt.

\paragraph{Intuition.}
If we perform a rough clustering based on our intuition about regions, we can see some trends: LLMs are more aligned with English-speaking countries and several Latin American and Central and Eastern European countries (e.g., the United States, Canada, Chile, the Czech Republic, Slovakia, and Uruguay). They are less aligned with several large Asian countries (e.g., China, India, and Indonesia).

\subsection{Alignment Difference Level}
\label{sec:alignment_difference_level}
We provide the relationship of different countries to all seven LLMs' opinion distribution and the average human opinion distribution alignment scores; the checklist is shown in Table~\ref{tab:compare_llms_humans}.

\begin{figure*}[t]
\centering
\includegraphics[width=0.65\linewidth]{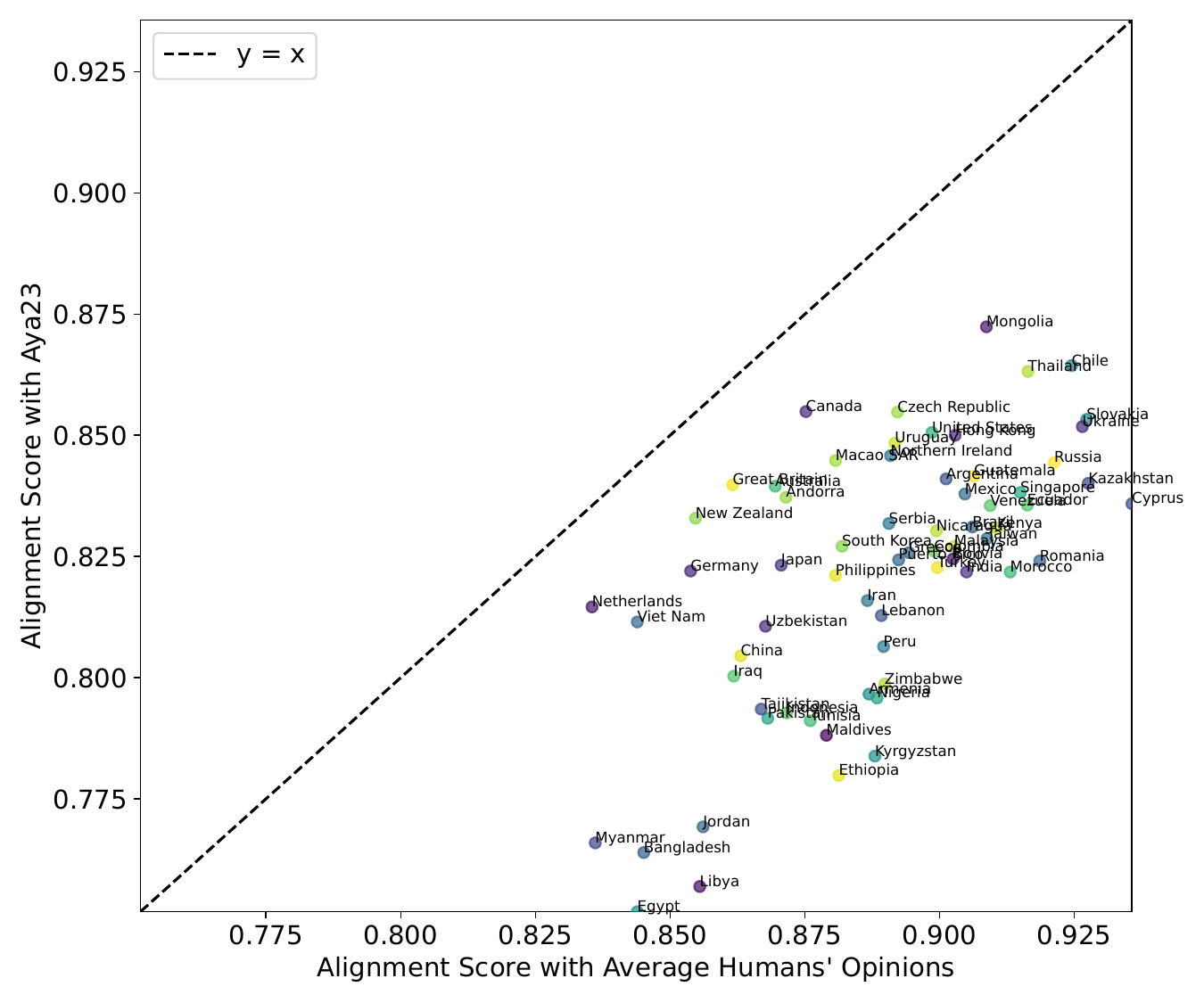}
\caption{The relationship of different countries to Aya23's opinion distribution and the average human opinion distribution alignment scores.}
\label{fig:compare_llms_humans_aya23}
\end{figure*}

\begin{figure*}[t]
\centering
\includegraphics[width=0.65\linewidth]{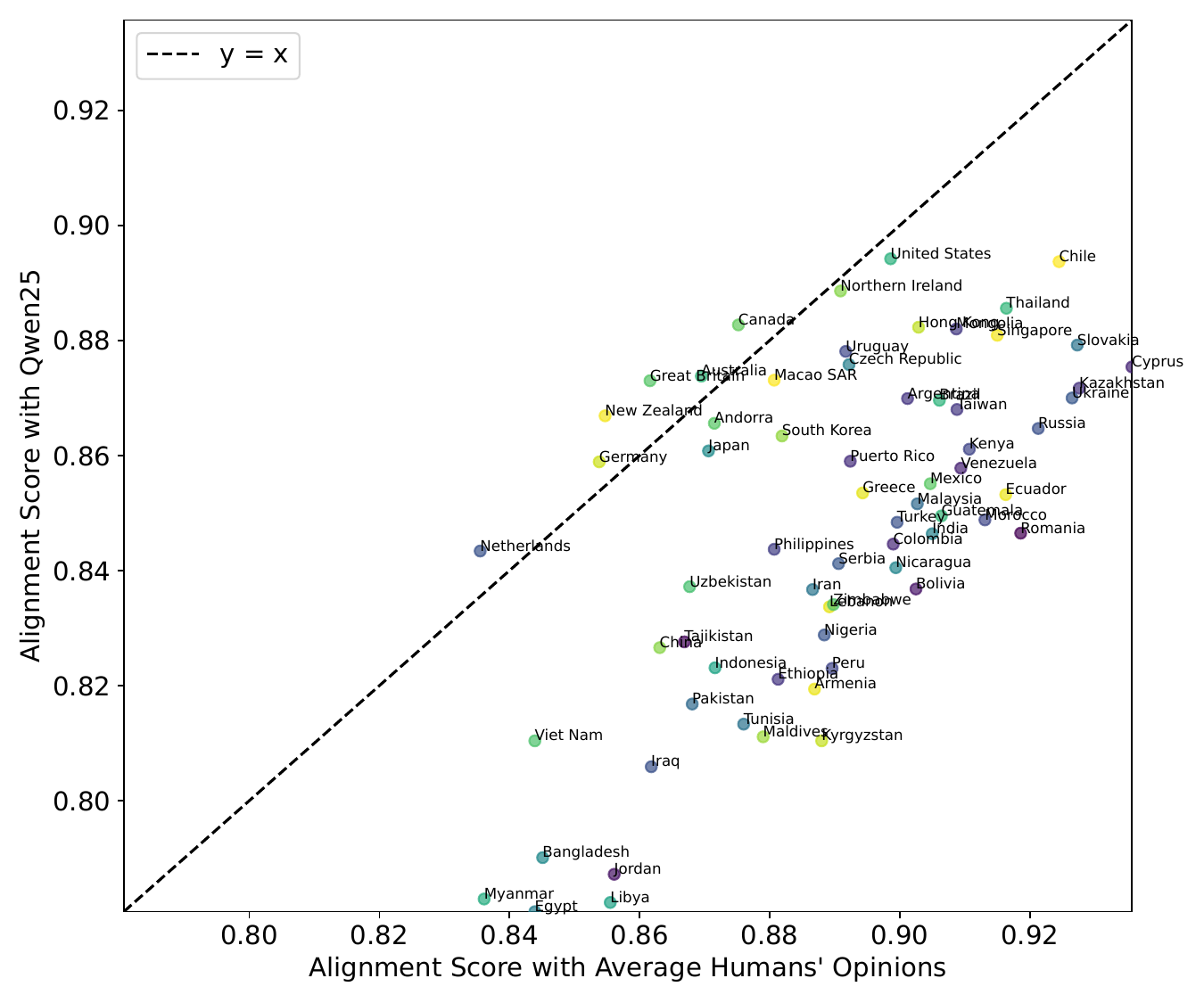}
\caption{The relationship of different countries to Qwen2.5's opinion distribution and the average human opinion distribution alignment scores.}
\label{fig:compare_llms_humans_qwen25}
\end{figure*}

\begin{figure*}[t]
\centering
\includegraphics[width=0.65\linewidth]{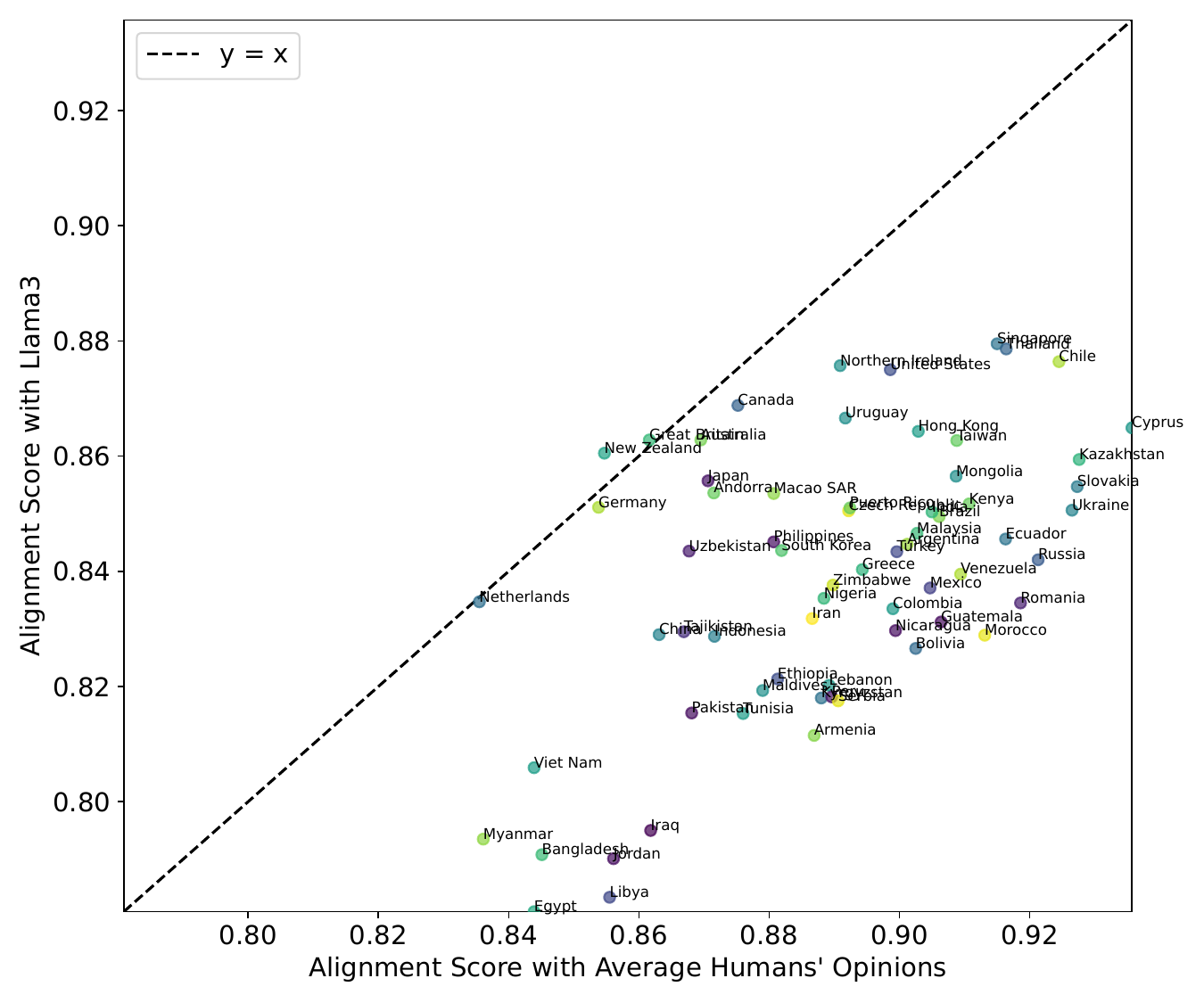}
\caption{The relationship of different countries to Llama3's opinion distribution and the average human opinion distribution alignment scores.}
\label{fig:compare_llms_humans_llama3}
\end{figure*}

\begin{figure*}[t]
\centering
\includegraphics[width=0.65\linewidth]{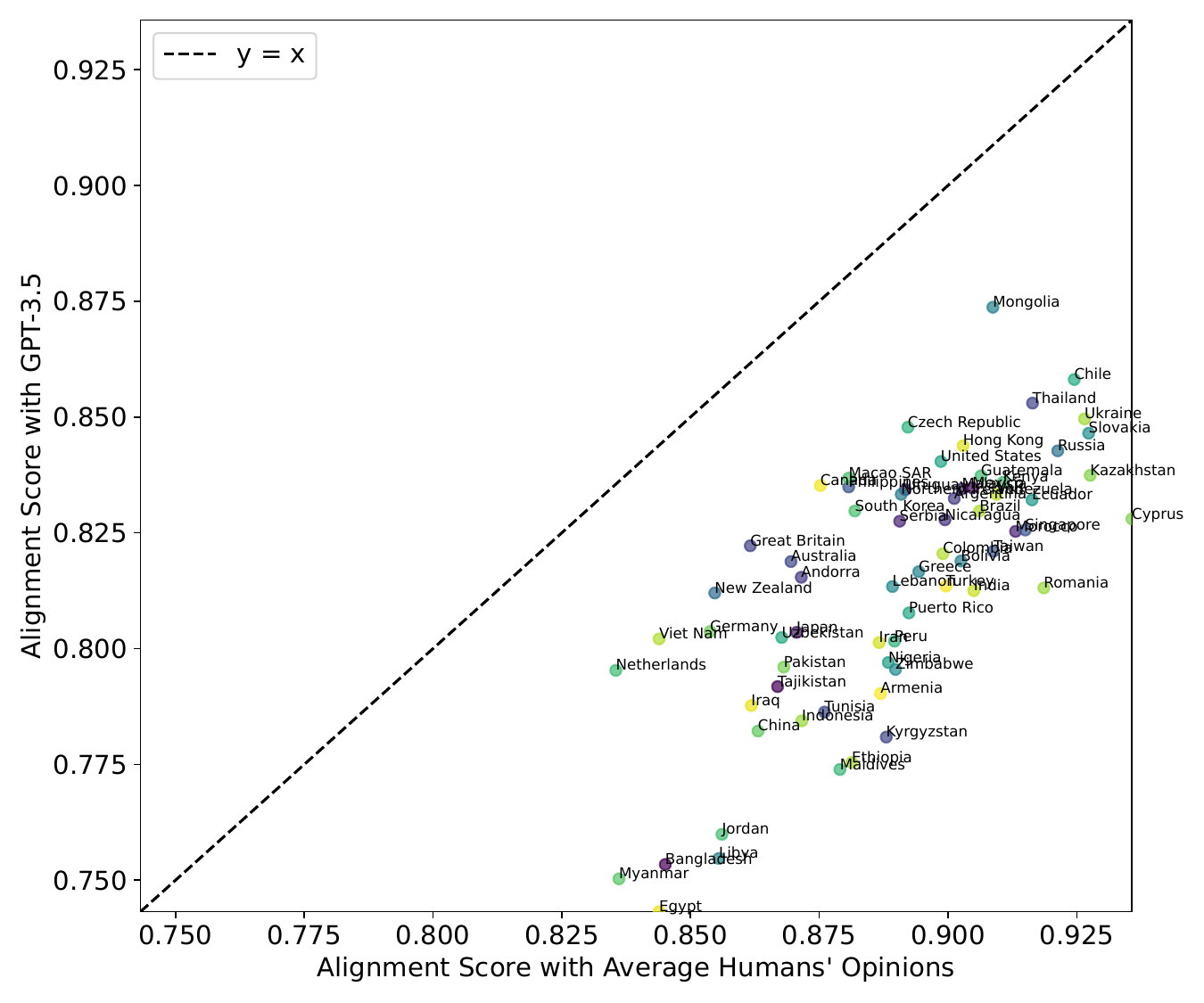}
\caption{The relationship of different countries to GPT-3.5's opinion distribution and the average human opinion distribution alignment scores.}
\label{fig:compare_llms_humans_gpt35}
\end{figure*}

\begin{figure*}[t]
\centering
\includegraphics[width=0.65\linewidth]{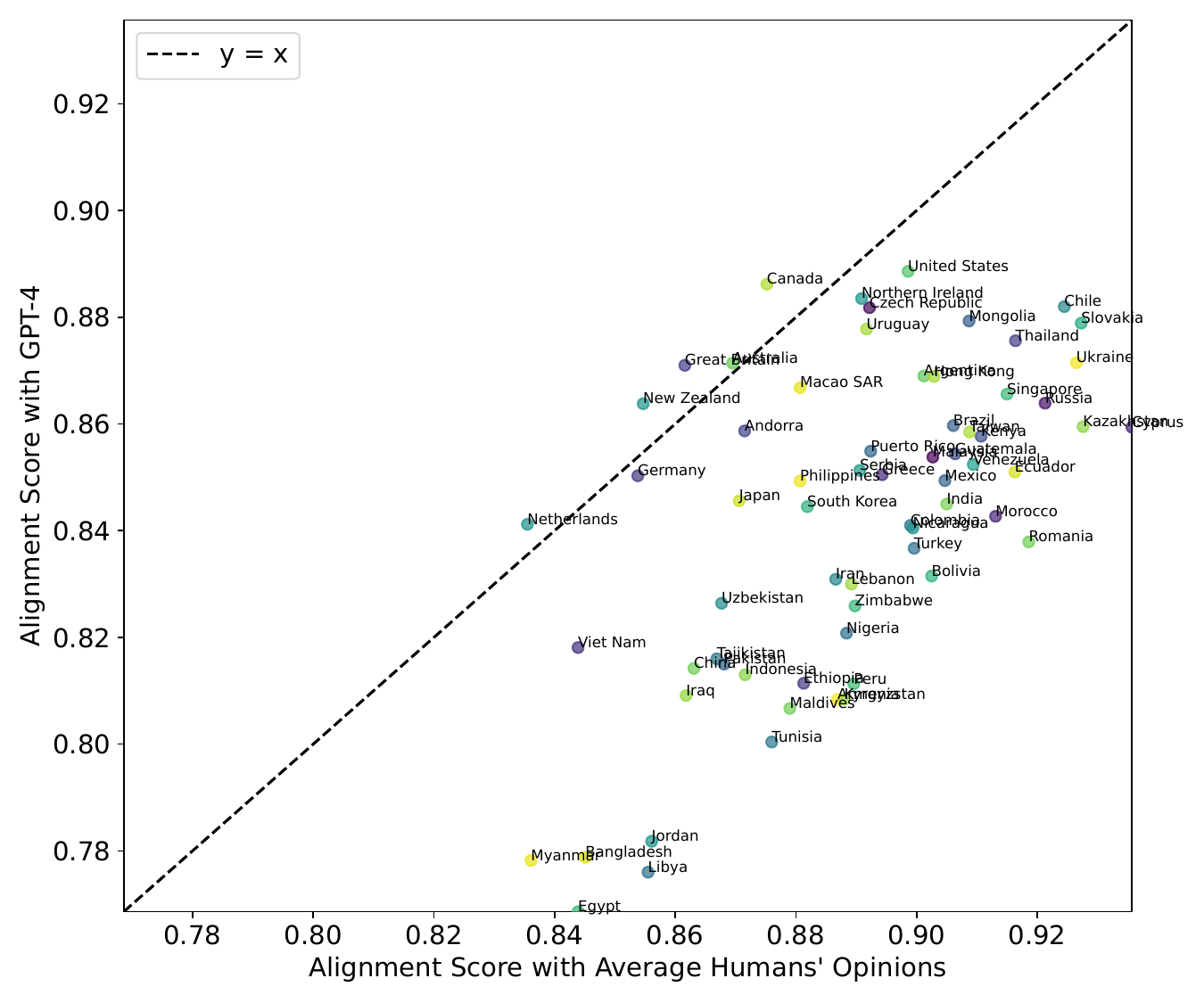}
\caption{The relationship of different countries to GPT-4's opinion distribution and the average human opinion distribution alignment scores.}
\label{fig:compare_llms_humans_gpt4}
\end{figure*}

\begin{figure*}[t]
\centering
\includegraphics[width=0.65\linewidth]{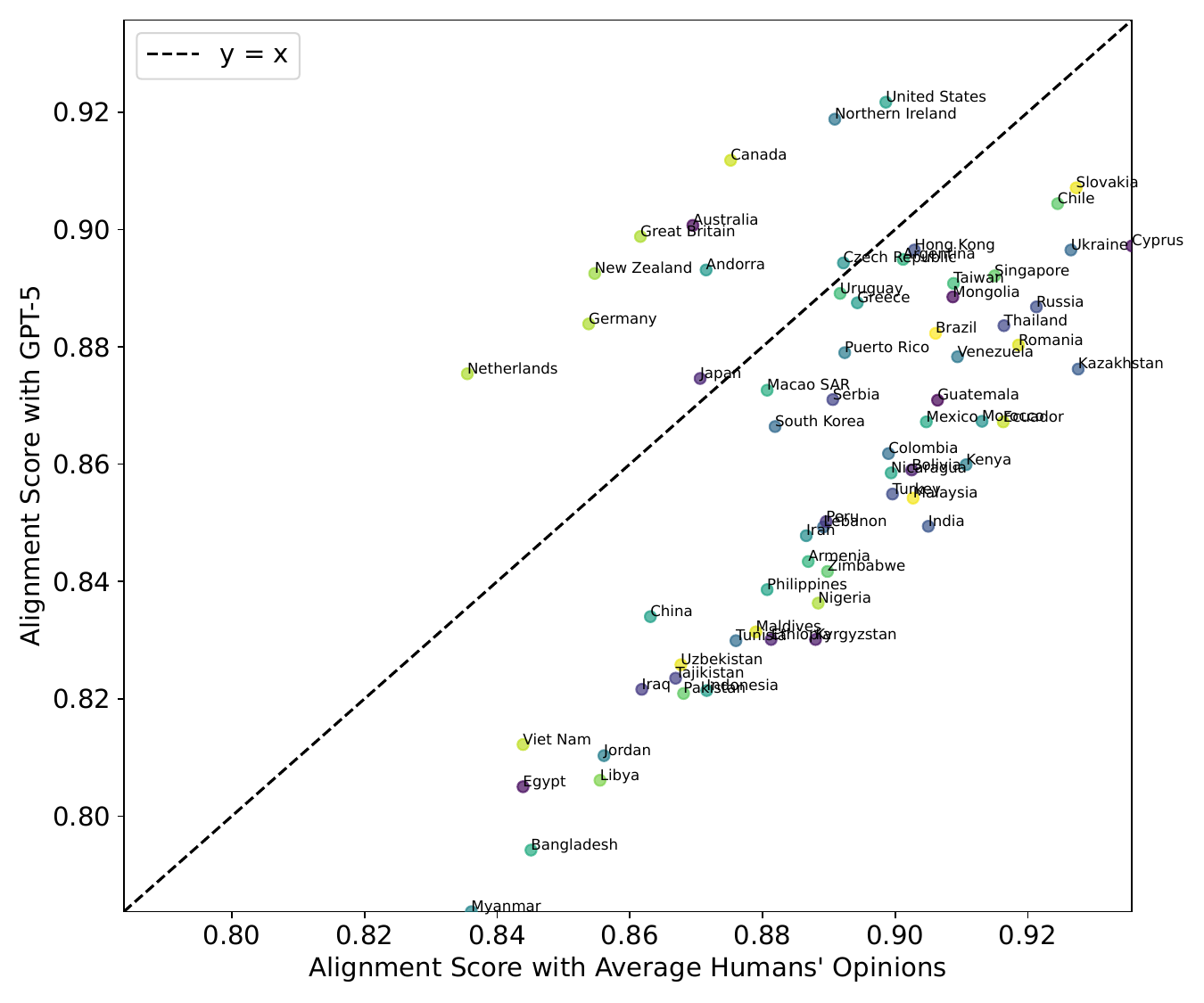}
\caption{The relationship of different countries to GPT-5's opinion distribution and the average human opinion distribution alignment scores.}
\label{fig:compare_llms_humans_gpt5}
\end{figure*}

\begin{figure*}[t]
\centering
\includegraphics[width=0.65\linewidth]{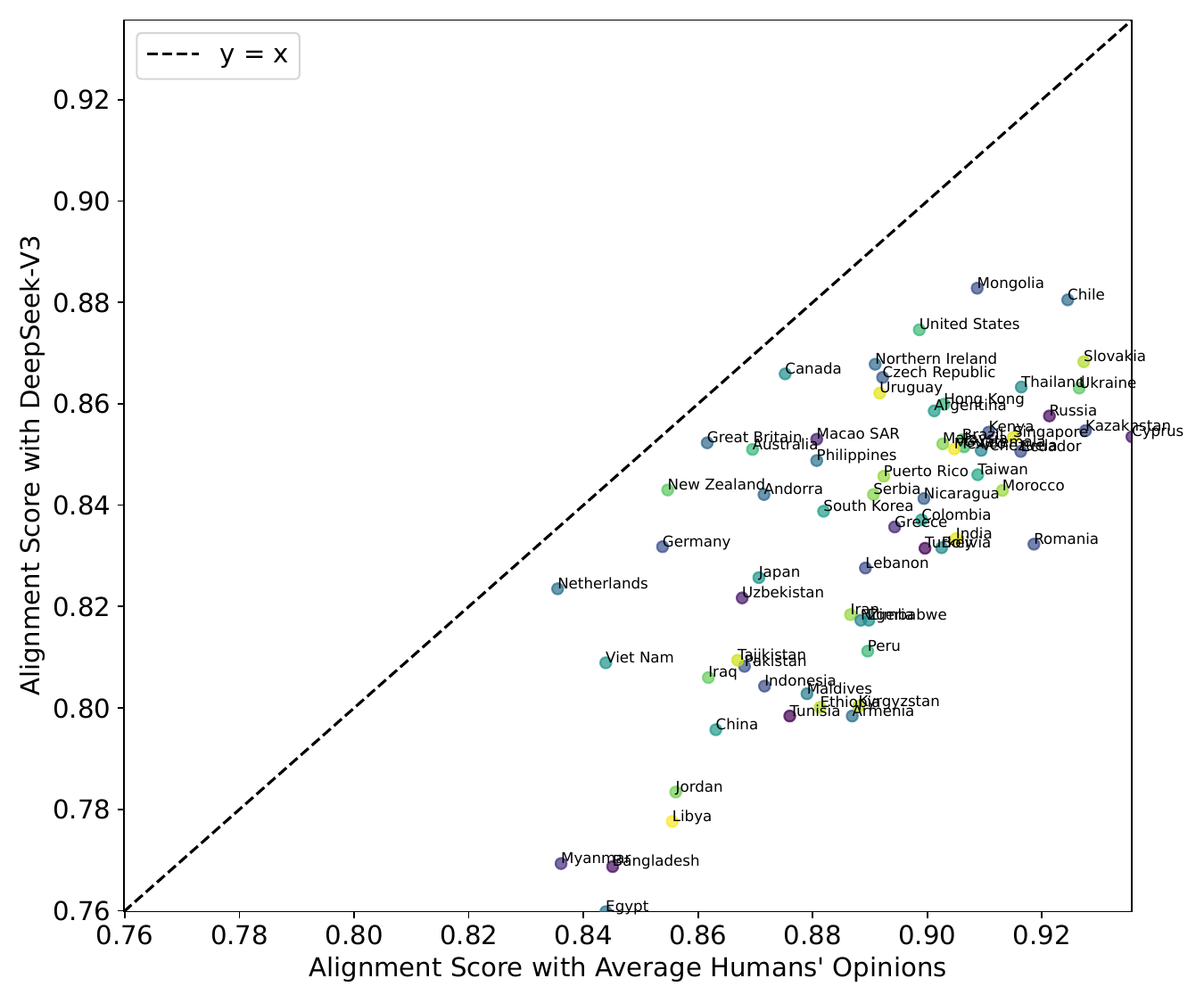}
\caption{The relationship of different countries to DeepSeek-V3's opinion distribution and the average human opinion distribution alignment scores.}
\label{fig:compare_llms_humans_dsv3}
\end{figure*}

\begin{figure*}[t]
\centering
\includegraphics[width=0.65\linewidth]{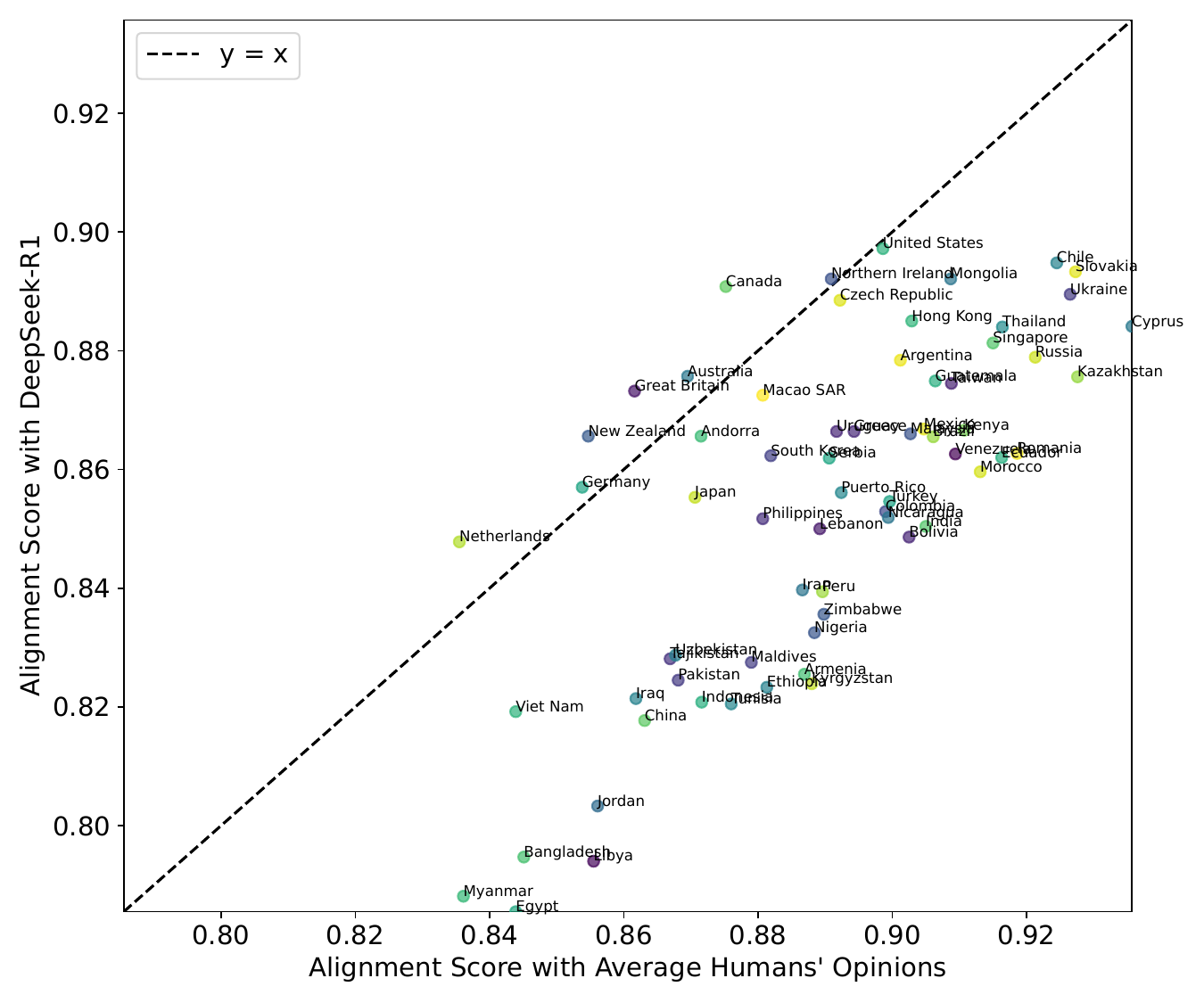}
\caption{The relationship of different countries to DeepSeek-R1's opinion distribution and the average human opinion distribution alignment scores.}
\label{fig:compare_llms_humans_dsr1}
\end{figure*}

\begin{table}[H]
    \centering
    \begin{tabular}{ll}
            \textbf{Models} &   \textbf{Figures}  \\
            \midrule
            Aya23 & Figure~\ref{fig:compare_llms_humans_aya23} \\
            Qwen2.5 & Figure~\ref{fig:compare_llms_humans_qwen25} \\
            Llama3 & Figure~\ref{fig:compare_llms_humans_llama3} \\
            GPT-3.5 & Figure~\ref{fig:compare_llms_humans_gpt35} \\
            GPT-4 & Figure~\ref{fig:compare_llms_humans_gpt4} \\
            GPT-5 & Figure~\ref{fig:compare_llms_humans_gpt5} \\
            DeepSeek-V3 & Figure~\ref{fig:compare_llms_humans_dsv3} \\
            DeepSeek-R1 & Figure~\ref{fig:compare_llms_humans_dsr1} \\
    \end{tabular}
    \caption{Checklist of relationship of different countries to all seven LLMs' opinion distribution and the average human opinion distribution alignment scores.}
    \label{tab:compare_llms_humans}
\end{table}

We find that Aya23, GPT-3.5, and DeepSeek-V3 do not achieve alignment scores higher than the average human alignment with any country.
Other LLMs align well with only a few countries, such as the United States and Canada, but show poor alignment with most countries.
This indicates that aligning LLMs with human opinions remains a considerable challenge. In addition, it should be noted that the alignment scores of LLMs with countries are positively correlated with the alignment scores of average human distributions with countries; that is, LLMs are more likely to align with countries that have more aligned average human distributions.

\section{More Results of RQ2}
\label{sec:more_results_of_rq2}

\begin{table*}[h]
    \centering
    \small
    % \tiny % todo: tiny may be not acceptable
    \scalebox{0.90}{
    \begin{tabular}{llllllllll}
    \toprule
    \textbf{Method} & \textbf{Aya23} & \textbf{Llama3} & \textbf{Qwen2.5} & \textbf{GPT-3.5} & \textbf{GPT-4} & \textbf{GPT-5} & \textbf{DS-V3} & \textbf{DS-R1} & \textbf{AVG.} \\
    
    \midrule
    \rowcolor{gray!10} \multicolumn{10}{c}{ \textit{Japan (Japanese)}} \\
    \midrule
    
    No Steering                          & 0.8238 & 0.8513 & 0.8596  &0.8015 & 0.8441 & 0.8744 & 0.8251 & 0.8541 & 0.8417 \\ 
    \ \ +\textit{Language Steering} (Ja.) & \textbf{0.8400} & \textbf{0.8516} & \textbf{0.8686} & \textbf{0.8141} & \textbf{0.8762}$^{**}$ & \textbf{0.8945}$^{*}$ & \textbf{0.8760}$^{***}$ & \textbf{0.8931}$^{***}$ & \textbf{0.8643}$^{***}$ \\ 

    \hline

    Persona Steering (En.)       & 0.8400 & \textbf{0.8566} & 0.8631 & 0.8035 & 0.8603 & 0.8884 & 0.8422 & 0.8680 & 0.8528$^{*}$ \\ 
    \ \ +\textit{Language Steering} (Ja.)  & \textbf{0.8444}$^{*}$ & 0.8540 & \textbf{0.8733} & \textbf{0.8287} & \textbf{0.8899}$^{***}$ & \textbf{0.9123}$^{***}$ & \textbf{0.8847}$^{***}$ & \textbf{0.8996}$^{***}$ & \textbf{0.8734}$^{***}$ \\ 

    \hline
    
    Few-shot Steering (En.)     & 0.8402 & \textbf{0.8618} & 0.8696 & \textbf{0.8327}$^{*}$ & 0.8757$^{**}$ & 0.8920 & 0.8810$^{***}$ & 0.8788$^{*}$ & 0.8665$^{***}$ \\ 
    \ \ +\textit{Language Steering} (Ja.) & \textbf{0.8696}$^{***}$ & 0.8508 & \textbf{0.8807}$^{*}$ & 0.8280 & \textbf{0.9027}$^{***}$ & \textbf{0.9053}$^{***}$ & \textbf{0.8932}$^{***}$ & \textbf{0.8964}$^{***}$ & \textbf{0.8783}$^{***}$ \\

    \midrule
    \rowcolor{gray!10} \multicolumn{10}{c}{ \textit{Korea (Korean)}} \\
    \midrule
    
    No Steering (En.)                   & \textbf{0.8244} & 0.8415 & \textbf{0.8632} & 0.8283 & 0.8445 & 0.8659 & 0.8378 & 0.8606 & 0.8458 \\ 
    \ \ +\textit{Language Steering} (Ko.)   & 0.8230 & \textbf{0.8427} & 0.8607 & \textbf{0.8379} & \textbf{0.8530} & \textbf{0.8737} & \textbf{0.8685}$^{*}$ & \textbf{0.8693} & \textbf{0.8536} \\ 
    
    \hline
    
    Persona Steering (En.)          & 0.8341 & 0.8357 & 0.8640 & 0.8256 & \textbf{0.8721}$^{*}$ & 0.8814$^{*}$ & 0.8747$^{**}$ & 0.8875$^{**}$ & 0.8594$^{***}$ \\ 
    \ \ +\textit{Language Steering} (Ko.)   & \textbf{0.8367} & \textbf{0.8479} & \textbf{0.8660} & \textbf{0.8354} & 0.8682$^{*}$ & \textbf{0.8844} & \textbf{0.8786}$^{***}$ & \textbf{0.8932}$^{**}$ & \textbf{0.8638}$^{***}$ \\ 
    
    \hline
    
    Few-shot Steering (En.)      & 0.8313 &0.8591 & 0.8646 & 0.8405 & 0.8700$^{*}$ & 0.8718 & 0.8831$^{***}$ & 0.8637 & 0.8605$^{***}$ \\ 
    \ \ +\textit{Language Steering} (Ko.) & \textbf{0.8531}$^{*}$ & \textbf{0.8749}$^{**}$ & \textbf{0.8790} & \textbf{0.8441} & \textbf{0.8857}$^{**}$ & \textbf{0.8960}$^{*}$ & \textbf{0.8953}$^{***}$ & \textbf{0.9019}$^{***}$ & \textbf{0.8788}$^{***}$ \\

    \midrule
    \rowcolor{gray!10} \multicolumn{10}{c}{ \textit{Russia (Russian)}} \\
    \midrule
    
    No Steering (En.)                       & \textbf{0.8421} & 0.8415 & 0.8644 & \textbf{0.8414} & 0.8640 & 0.8862 & 0.8569 & 0.8780 & 0.8593 \\ 
    \ \ +\textit{Language Steering} (Ru.)      & 0.8347 & \textbf{0.8682}$^{*}$ & \textbf{0.8776} & 0.8408 & \textbf{0.8977}$^{**}$ & \textbf{0.9103}$^{**}$ & \textbf{0.9001}$^{***}$ & \textbf{0.8880} & \textbf{0.8772}$^{***}$ \\ 

    \hline
    
    Persona Steering (En.)            & 0.8495 & 0.8513 & 0.8854 & \textbf{0.8578} & 0.8953$^{**}$ & \textbf{0.9180}$^{***}$ & 0.8978$^{***}$ & 0.9018$^{*}$ & 0.8821$^{***}$ \\ 
    \ \ +\textit{Language Steering} (Ru.)    & \textbf{0.8668}$^{*}$ & \textbf{0.8729}$^{**}$ & \textbf{0.8895}$^{*}$ & 0.8344 & \textbf{0.9002}$^{***}$ & 0.8976 & \textbf{0.9129}$^{***}$ & \textbf{0.9084}$^{***}$ & \textbf{0.8853}$^{***}$ \\ 
    
    \hline
    
    Few-shot Steering (En.)       & 0.8562 & 0.8801$^{***}$ & 0.8865 & \textbf{0.8700}$^{*}$ & 0.8936$^{**}$ & 0.9172$^{***}$ & 0.9058$^{***}$ & 0.8939 & 0.8879$^{***}$ \\ 
    \ \ +\textit{Language Steering} (Ru.)  & \textbf{0.8699}$^{*}$ & \textbf{0.8874}$^{***}$ & \textbf{0.8947}$^{**}$ & 0.8457 & \textbf{0.9095}$^{***}$ & \textbf{0.9247}$^{***}$ & \textbf{0.9084}$^{***}$ & \textbf{0.8975}$^{*}$ & \textbf{0.8922}$^{***}$ \\ 

    \midrule
    \rowcolor{gray!10} \multicolumn{10}{c}{ \textit{Viet Nam (Vietnamese)}} \\
    \midrule
    
    No Steering (En.)                  & 0.8105 & 0.7992 & 0.8094 & 0.8019 & 0.8179 & 0.8111 & 0.8078 & 0.8164 & 0.8093 \\ 
    \ \ +\textit{Language Steering} (Vi.)   & \textbf{0.8129} & \textbf{0.8364}$^{**}$ & \textbf{0.8487}$^{**}$ & \textbf{0.8062} & \textbf{0.8523}$^{**}$ & \textbf{0.8407}$^{*}$ & \textbf{0.8489}$^{***}$ & \textbf{0.8496}$^{**}$ & \textbf{0.8370}$^{***}$ \\ 
    
    \hline
    
    Persona Steering (En.)         & 0.8070 & 0.8078 & 0.7987 & 0.7975 & 0.8235 & 0.8337 & 0.8227 & 0.8253 & 0.8145 \\ 
    \ \ +\textit{Language Steering} (Vi.)   & \textbf{0.8100} & \textbf{0.8382}$^{**}$ & \textbf{0.8408}$^{*}$ & \textbf{0.8114} & \textbf{0.8612}$^{***}$ & \textbf{0.8544}$^{***}$ & \textbf{0.8481}$^{**}$ & \textbf{0.8525}$^{**}$ & \textbf{0.8396}$^{***}$ \\ 
    \hline
    Few-shot Steering (En.)        & 0.7961 & 0.8498$^{***}$ & 0.8190 & 0.8218 & 0.8470$^{*}$ & 0.8450$^{*}$ & 0.8444$^{*}$ & 0.8396 & 0.8328$^{***}$ \\ 
    \ \ +\textit{Language Steering} (Vi.)   & \textbf{0.8443}$^{**}$ & \textbf{0.8668}$^{***}$ & \textbf{0.8718}$^{***}$ & \textbf{0.8411}$^{**}$ & \textbf{0.8808}$^{***}$ & \textbf{0.8713}$^{***}$ & \textbf{0.8854}$^{***}$ & \textbf{0.8674}$^{***}$ & \textbf{0.8661}$^{***}$ \\ 

    \midrule
    \rowcolor{gray!10} \multicolumn{10}{c}{ \textit{Brazil (Portuguese)}} \\
    \midrule 
  
    No Steering (En.)                  & 0.8294 & 0.8455 & 0.8693 & 0.8288 & 0.8606 & \textbf{0.8834} & 0.8529 & 0.8655 & 0.8544 \\ 
    \ \ +\textit{Language Steering} (Pt.)   & \textbf{0.8450} & \textbf{0.8557} & \textbf{0.8791} & \textbf{0.8453} & \textbf{0.8914}$^{***}$ & 0.8753 & \textbf{0.8882}$^{***}$ & \textbf{0.8828} & \textbf{0.8704}$^{***}$ \\ 
    
    \hline
    
    Persona Steering (En.)         & \textbf{0.8583}$^{*}$ & 0.8380 & 0.8820 & 0.8422 & 0.8855$^{**}$ & \textbf{0.8983} & 0.8792$^{**}$ & 0.8815 & 0.8706$^{***}$ \\ 
    \ \ +\textit{Language Steering} (Pt.)   & 0.8552 & \textbf{0.8680}$^{*}$ & \textbf{0.8885}$^{*}$ & \textbf{0.8489} & \textbf{0.8992}$^{***}$ & 0.8878 & \textbf{0.8945}$^{***}$ & \textbf{0.8872}$^{*}$ & \textbf{0.8787}$^{***}$ \\ 
    \hline
    Few-shot Steering (En.)       & 0.8518 & \textbf{0.8777}$^{**}$ & 0.8842 & 0.8696$^{***}$ & 0.8879$^{**}$ & 0.9047$^{*}$ & 0.8964$^{***}$ & 0.8805 & 0.8816$^{***}$ \\ 
    \ \ +\textit{Language Steering} (Pt.)   & \textbf{0.8569}$^{*}$ & 0.8755$^{**}$ & \textbf{0.8869} & \textbf{0.8720}$^{***}$ & \textbf{0.8995}$^{***}$ & \textbf{0.9049}$^{**}$ & \textbf{0.8987}$^{***}$ & \textbf{0.8939}$^{*}$ & \textbf{0.8860}$^{***}$ \\ 

    \midrule
    \rowcolor{gray!10} \multicolumn{10}{c}{ \textit{Argentina (Spanish)}} \\
    \midrule 
     
    No Steering (En.)                   & \textbf{0.8392} & 0.8397 & 0.8698 & \textbf{0.8310} & 0.8694 & 0.8943 & 0.8579 & 0.8766 & 0.8597 \\ 
    \ \ +\textit{Language Steering} (Es.)   & 0.8349 & \textbf{0.8518} & \textbf{0.8764} & 0.8138 & \textbf{0.8890}$^{*}$ & \textbf{0.8953} &\textbf{0.8868}$^{**}$ & \textbf{0.8912} & \textbf{0.8674} \\ 
    \hline
    Persona Steering (En.)         & 0.8294 & 0.8429 & 0.8774 & 0.8248 & 0.8869 & \textbf{0.9101} & 0.8804$^{*}$ & 0.8929 & 0.8681$^{*}$ \\ 
    \ \ +\textit{Language Steering} (Es.)   & \textbf{0.8540} & \textbf{0.8628} & \textbf{0.8874}$^{*}$ & \textbf{0.8303} & \textbf{0.8896}$^{*}$ & 0.9063 & \textbf{0.8921}$^{***}$ & \textbf{0.8980}$^{*}$ & \textbf{0.8776}$^{***}$ \\ 
    \hline
    Few-shot Steering (En.)        & \textbf{0.8612} & 0.8627 & \textbf{0.8984}$^{*}$ & \textbf{0.8483}$^{**}$ & \textbf{0.9021}$^{***}$ & \textbf{0.9153}$^{**}$ & 0.9052$^{***}$ & 0.8947 & \textbf{0.8860}$^{***}$ \\ 
    \ \ +\textit{Language Steering} (Es.)   & 0.8546 & \textbf{0.8684} & 0.8983$^{**}$ & 0.8267$^{**}$ & 0.8999$^{**}$ & 0.9080 & \textbf{0.9099}$^{***}$ & \textbf{0.8985}$^{*}$ & 0.8830$^{***}$ \\ 

    \midrule
    \rowcolor{gray!10} \multicolumn{10}{c}{ \textit{Chile (Spanish)}} \\
    \midrule 
    
    No Steering (En.)                       & \textbf{0.8627} & 0.8728 & 0.8936 & \textbf{0.8570} & 0.8830 & \textbf{0.9051} & 0.8798 & 0.8937 & 0.8810 \\ 
    \ \ +\textit{Language Steering} (Es.)       & 0.8508 & \textbf{0.8824} & \textbf{0.8985} & 0.8444 & \textbf{0.9074}$^{**}$ & 0.9044 & \textbf{0.9076}$^{**}$ & \textbf{0.9033} & \textbf{0.8874} \\ 
    
    \hline
    
    Persona Steering (En.)               & 0.8561 & 0.8836 & 0.9040 & 0.8486 & 0.9090$^{**}$ & \textbf{0.9152} & 0.9054$^{**}$ & \textbf{0.9091} & 0.8914$^{**}$ \\ 
    \ \ +\textit{Language Steering} (Es.)         & \textbf{0.8634} & \textbf{0.8929}$^{*}$ & \textbf{0.9047} & \textbf{0.8556} & \textbf{0.9120}$^{***}$ & 0.9132 & \textbf{0.9111}$^{***}$ & 0.9012 & \textbf{0.8943}$^{***}$ \\ 
    
    \hline
    
    Few-shot Steering (En.)               & \textbf{0.8757} & 0.8987$^{**}$ & \textbf{0.9075} & \textbf{0.8607} & 0.9141$^{***}$ & \textbf{0.9230}$^{*}$ & 0.9150$^{***}$ & 0.8989 & 0.8992$^{***}$ \\ 
    \ \ +\textit{Language Steering} (Es.)           & 0.8660 & \textbf{0.9019}$^{***}$ & 0.9020 & 0.8569 & \textbf{0.9177}$^{***}$ & 0.9210$^{*}$ & \textbf{0.9198}$^{***}$ & \textbf{0.9116}$^{*}$ & \textbf{0.8996}$^{***}$ \\ 

    \midrule
    \rowcolor{gray!10} \multicolumn{10}{c}{ \textit{Uruguay (Spanish)}} \\
    \midrule 
    
    No Steering (En.)                       & \textbf{0.8475} & \textbf{0.8607} & \textbf{0.8766} & \textbf{0.8322} & 0.8767 & \textbf{0.8893} & 0.8595 & 0.8639 & 0.8633 \\ 
    \ \ +\textit{Language Steering} (Es.)        & 0.8389 & 0.8592 & 0.8744 & 0.8222 & \textbf{0.8858} & 0.8831 & \textbf{0.8769} & \textbf{0.8711} & \textbf{0.8640} \\ 
    
    \hline
    
    Persona Steering (En.)             & 0.8466 & 0.8584 & 0.8797 & 0.8219 & 0.8742 & 0.8946 & \textbf{0.8810}$^{*}$ & \textbf{0.8900}$^{**}$ & 0.8683 \\ 
    \ \ +\textit{Language Steering} (Es.)       & \textbf{0.8534} & \textbf{0.8648} & \textbf{0.8802} & \textbf{0.8345} & \textbf{0.8823} & \textbf{0.8950} & 0.8797$^{*}$ & 0.8898$^{**}$ & \textbf{0.8725}$^{**}$ \\ 
    
    \hline
    
    Few-shot Steering (En.)           & \textbf{0.8564} & 0.8642 & 0.8849 & 0.8267 & 0.8853 & \textbf{0.9050} & 0.8850$^{**}$ & 0.8804 & 0.8735$^{**}$ \\ 
    \ \ +\textit{Language Steering} (Es.)       & 0.8513 & \textbf{0.8709} & \textbf{0.8899} & \textbf{0.8286} & \textbf{0.8921} & 0.8989 & \textbf{0.8854}$^{**}$ & \textbf{0.8923}$^{***}$ & \textbf{0.8762}$^{***}$ \\ 
    
    \bottomrule
    \end{tabular}
    }
    \caption{The alignment scores between Japan, Korea, Russia, Viet Nam, Brazil, Argentina, Chile, and Uruguay and LLMs under different steering methods. 
    The content in ``()'' is to indicate the language being used, where ``En.'' denotes English, ``Ja.'' denotes Japanese, ``Ko.'' denotes Korean, ``Ru.'' denotes Russian, `Vi.'' denotes Vietnamese, `Pt.'' denotes Portuguese, and `Es.'' denotes Spanish. In all settings, our input always keeps single language.
    Moreover, the significance is assessed using the $t$-test: * denotes $p$-value $<$ 0.05, ** denotes $p$-value $<$ 0.01, and *** denotes $p$-value $<$ 0.001.}
    \label{tab:all_steerability_results}
\end{table*}

In addition to the Chinese and German in the main text, we provide the experimental results of Japanese, Korean, Russian, Vietnamese, Portuguese, and Spanish in Table~\ref{tab:all_steerability_results}.
The experimental results show that language steering is more effective when a language is used by a single country. However, when a language such as Spanish is used by many countries, language steering may cause LLMs to express the average opinion of Spanish-speaking countries rather than the opinion of a specific country.
This result suggests that when steering LLMs to emulate the opinions of these countries, features other than language should also be considered.

\section{More Results of RQ3}
\label{sec:more_results_of_rq3}
We use wave 7's survey data to filter countries for each LLM, and Table~\ref{tab:filtered_country_list} provides a list of countries filtered for each LLM. In the case of the threshold $\tau=0.02$, Aya-23-35B, GPT-3.5-Turbo, and DeepSeek-V3 cover less than five countries, so we do not consider these models. 
Table~\ref{fig:human_performance_all} shows the trend of the average alignment scores of other LLMs with countries across waves.

\begin{table*}[t]
    \centering
    \begin{tabular}{p{3cm}p{8cm}}
            \textbf{Models} &   \textbf{Countries (or Regions)}  \\
            \midrule
            Qwen2.5 & \textit{Australia, Germany, Japan, Netherlands, New Zealand, South Korea, United States, Uruguay} \\
            Llama3 & \textit{Australia, Germany, Japan, Netherlands, New Zealand} \\
            GPT-4 & \textit{Australia, Germany, Netherlands, New Zealand, United States, Uruguay} \\
            GPT-5 & \textit{Argentina, Hong Kong, Japan, South Korea, Taiwan, Uruguay} \\
            DeepSeek-R1 & \textit{Australia, Germany, Hong Kong, Japan, Netherlands, New Zealand, United States} \\
    \end{tabular}
    \caption{Filtered countries for each LLM.}
    \label{tab:filtered_country_list}
\end{table*}

\begin{figure}[h]
\centering
\includegraphics[width=\columnwidth]{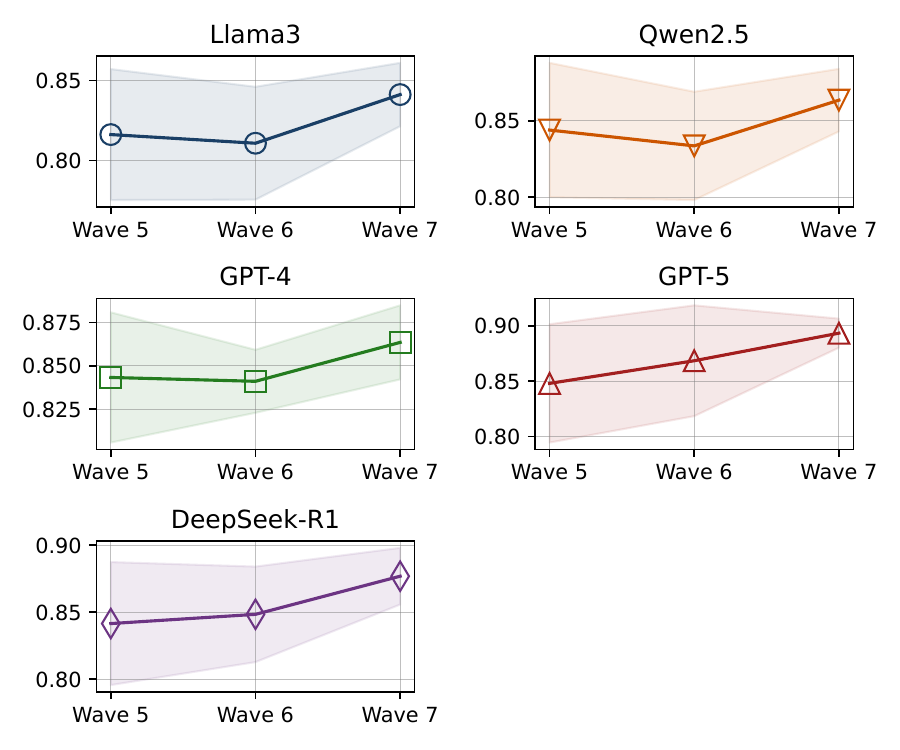}
\caption{The trend of the average alignment scores of LLMs with countries across waves. The sahed area indicates the standard deviation of the alignment scores at the current wave.}
\label{fig:human_performance_all}
\end{figure}

\section{More Results of Discussion}
\subsection{Alignment of Human Opinions across Countries}
\label{sec:alignment_of_human_opinions_across_countries}
We provide a full version of the alignment score heatmap between countries and LLMs in Figure~\ref{fig:human_performance_all}.

\begin{figure*}[t]
\centering
\includegraphics[width=\linewidth]{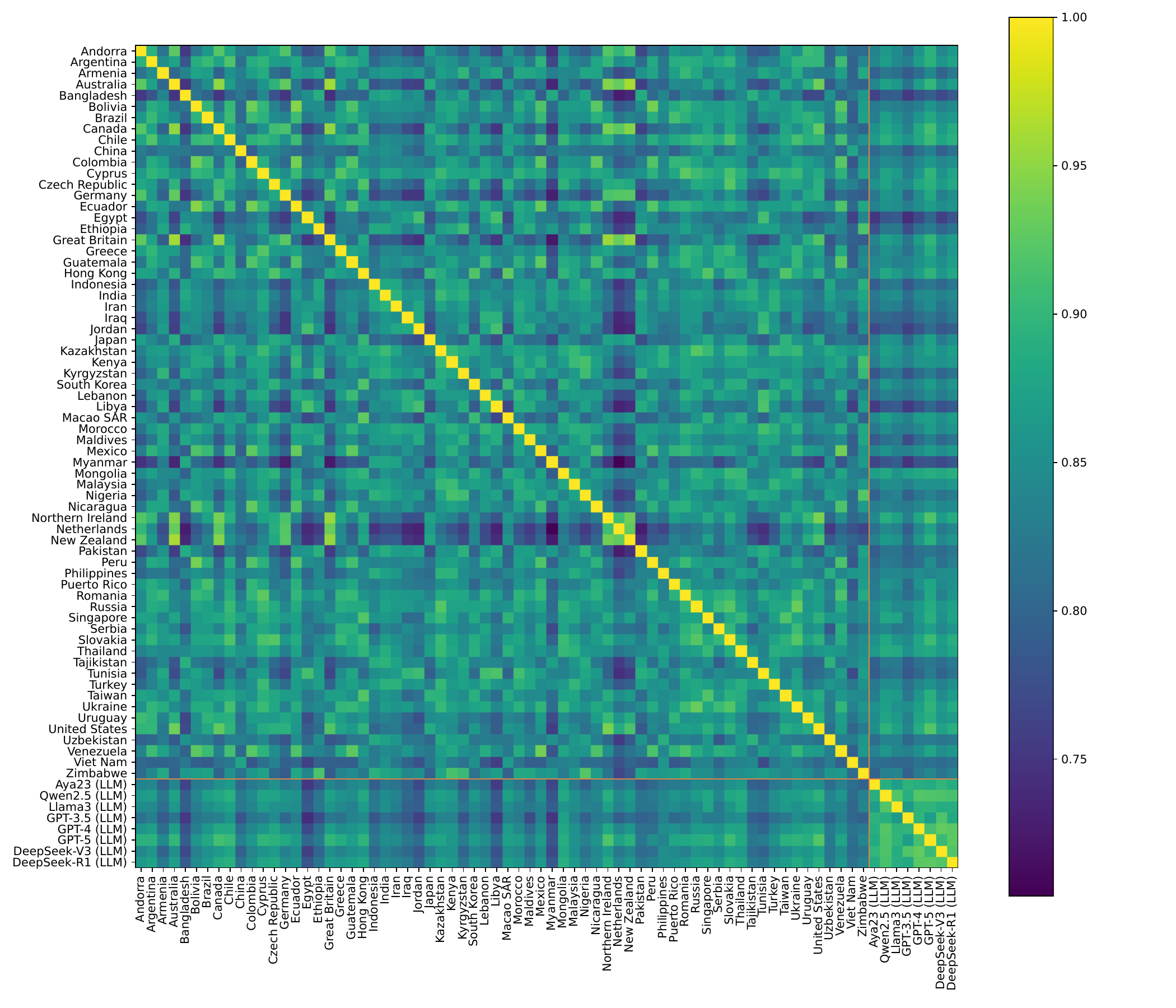}
\caption{A full version of the alignment score heatmap between countries and LLMs. The source code for plotting this figure is located in the \texttt{show\_results} folder.}
\label{fig:human_performance_all}
\end{figure*}

Individual columns and rows appear as ``darker horizontal and vertical stripes'': this means that some countries are less aligned with most other countries (systematic outliers). There are several noticeable dark stripes in the figure, such as Northern Ireland, the Netherlands, and New Zealand. However, these countries have high alignment scores with each other and with Australia, Brazil, and Great Britain.
The alignment scores between LLMs are relatively high, indicating a certain degree of homogeneity between the LLMs. This may be because they have similar training corpora~\citet{bommasani2022picking}. We also find that LLMs consistently exhibit poor alignment with certain countries.
For example, Bangladesh, Egypt, Iraq, Libya, and Myanmar, etc. These findings supplement the conclusions in the main text.

\subsection{Internal Consistency of LLMs}
\label{sec:internal_consistency_of_llms}

\begin{table*}[t]
    \centering
        \begin{tabular}{p{1.5cm}p{5cm}p{4cm}}
        \textbf{Topic} & \textbf{Question} & \textbf{Options}\\
        \midrule

        \multirow{13}{1.5cm}{Gender Fairness} 
        & Q29: Do you agree that, on the whole, men make better political leaders than women do? & 1. Strongly agree 2. Agree 3. Disagree 4. Strongly disagree\\
        & Q30: Do you agree that a university education is more important for a boy than for a girl? & 1. Strongly agree 2. Agree 3. Disagree 4. Strongly disagree\\
        & Q31: Do you agree that, on the whole, men make better business executives than women do? & 1. Strongly agree 2. Agree 3. Disagree 4. Strongly disagree\\
        & Q33: Do you agree that when jobs are scarce, men should have more right to a job than women? & 1. Strongly agree 2. Agree 3. Neither agree nor disagree 4. Disagree 5. Strongly disagree\\

        \midrule
        
        \multirow{5}{1.5cm}{Atheism}  & Q165: Do you believe in God? & 1. Yes 2. No \\
        & Q166: Do you believe in life after death? &  1. Yes 2. No \\
        & Q167: Do you believe in hell? &  1. Yes 2. No \\
        & Q168: Do you believe in heaven? &  1. Yes 2. No \\
    
        \midrule
        
        \multirow{7}{1.5cm}{Democracy}  & Q243: How essential do you think it is as a characteristic of democracy that people choose their leaders in free elections? & 1. Not an essential characteristic of democracy 10. An essential characteristic of democracy \\
        & Q250: How important is it for you to live in a country that is governed democratically? & 1. Not at all important 10. Absolutely important \\
        
        \end{tabular}
    
    \caption{The questions we selected for studying the internal consistency of LLMs.}
    \label{tab:consistency_examples}
\end{table*}

We provide a list of questions we selected for studying the internal consistency of LLMs in Table~\ref{tab:consistency_examples}.

\subsection{Sensitivity of LLMs}
\label{sec:sensitivity_of_llms}
Inspired by~\citet{santurkar2023whose}, we test the prompt sensitivity of the LLMs used in our experiments against the following factors:
\begin{itemize}
    \item The \textbf{order} in which the question options are presented to the model. The file \texttt{dataset/questions/WV7\_shuffle.jsonl} contains data with the order of options shuffled.
    \item The few-shot examples (the \textbf{number} of the few-shot examples and the \textbf{probability distribution} of the few-shot examples). The file \texttt{inputs/sensitivity/number.txt} is the examples that change the number of few-shot examples from 5 to 3. The file \texttt{inputs/sensitivity/prob.txt} is the examples that change the probability distribution of few-shot examples.
\end{itemize}

All results are listed under the \texttt{results/Discussion/sensitivity} folder.

\subsection{The Impact of Model Size}
\label{sec:g4}

\begin{figure*}[h]
\centering
\includegraphics[width=0.65\linewidth]{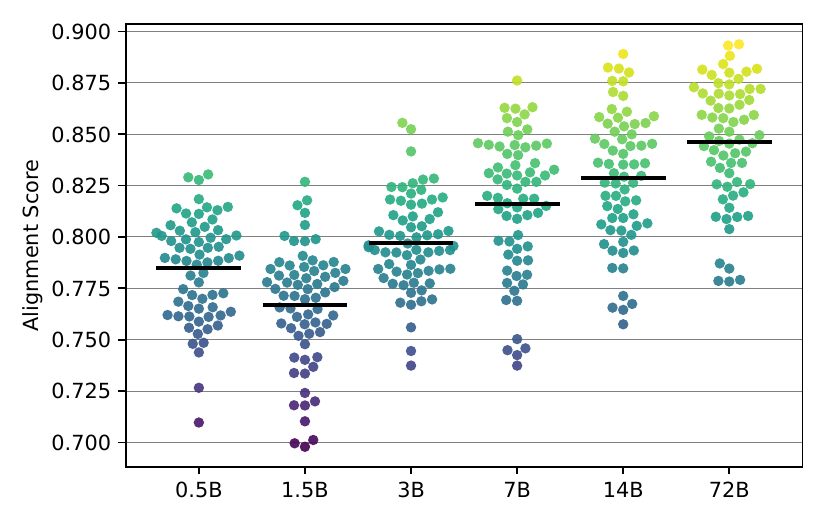}
\caption{The alignment score of Qwen 2.5 models of different sizes with each country, where each point represents a country and the black line represents the average alignment score of all countries.}
\label{fig:compare_llms_size}
\end{figure*}

To investigate the impact of different model sizes on experimental results, we evaluated 5 smaller Qwen2.5 (0.5B/1.5B/3B/7B/14B) models.
Figure~\ref{fig:compare_llms_size} shows the alignment scores of the Qwen2.5 model in different sizes for each country.
Although previous studies have shown that model alignment is not directly related to model size~\cite{liu-chu-2025-llms}. 
In Figure~\ref{fig:compare_llms_size}, we can observe that, except for Qwen2.5-1.5B-Instruct, the alignment score exhibits a clear positive correlation with model size, confirming that larger models more accurately capture the distribution of human values. 
We tests the language steering performance on Chinese and German across different size Qwen models. The experimental results are shown in Table~\ref{tab:slm_steerability_results}. Language steering also performed more effectively on larger models.

\begin{table*}[h]
    \centering
    \begin{tabular}{llllllll}
    \toprule
    \textbf{Method} & \textbf{0.5B} & \textbf{1.5B} & \textbf{3B} & \textbf{7B} & \textbf{14B} & \textbf{72B} & \textbf{AVG.} \\
    
    \midrule
    \rowcolor{gray!10} \multicolumn{8}{c}{ \textit{China}} \\
    \midrule
    
    No Steering (En.) & \textbf{0.7193} & \textbf{0.7569} & 0.7614 & 0.7960 & 0.8156 & 0.8273 & 0.7794 \\ 
    \ \ +\textit{Language Steering} (Zh.) & 0.7059 & 0.7427 & \textbf{0.7723} & \textbf{0.8173} & \textbf{0.8428} & \textbf{0.8439} & \textbf{0.7875} \\ 
    
    \hline 
    
    Persona Steering (En.)  & \textbf{0.7211} & \textbf{0.7627} & 0.7569 & 0.7990 & 0.8249 & 0.8316 & 0.7827 \\ 
    \ \ +\textit{Language Steering} (Zh.) & 0.7024 & 0.7560 & \textbf{0.7736} & \textbf{0.8183} & \textbf{0.8586}$^{**}$ & \textbf{0.8629}$^{*}$ & \textbf{0.7953}$^{*}$ \\ 
    
    \hline
    
    Few-shot Steering (En.) & 0.7341 & \textbf{0.7808} & \textbf{0.8072}$^{**}$ & 0.8143 & 0.8493 & 0.8596 & 0.8076$^{***}$ \\ 
    \ \ +\textit{Language Steering} (Zh.) & \textbf{0.7346} & 0.7751 & 0.7934 & \textbf{0.8357}$^{*}$ & \textbf{0.8754}$^{***}$ & \textbf{0.8838}$^{***}$ & \textbf{0.8163}$^{***}$ \\

    \midrule
    \rowcolor{gray!10} \multicolumn{8}{c}{ \textit{Germany}} \\
    \midrule
    
    No Steering (En.) & \textbf{0.7356} & \textbf{0.7680} & 0.7398 & \textbf{0.8227} & \textbf{0.8597} & 0.8541 & 0.7967\\ 
    \ \ +\textit{Language Steering} (De.) & 0.7331 & 0.7620 & \textbf{0.7761} & 0.8022 & 0.8469 & \textbf{0.8815}$^{**}$ & \textbf{0.8803} \\ 
    
    \hline
    
    Persona Steering (En.) & 0.7416 & \textbf{0.7757} & 0.7418 & \textbf{0.8244} & \textbf{0.8672} & 0.8596 & \textbf{0.8217} \\  
    \ \ +\textit{Language Steering} (De.) & \textbf{0.7431} & 0.7630 & \textbf{0.7898}$^{**}$ & 0.8086 & 0.8552 & \textbf{0.8904}$^{***}$ & 0.8084 \\  
    
    \hline
    
    Few-shot Steering (En.) & 0.7189 & 0.7531 & 0.7730 & 0.8200 & 0.8534 & 0.8619 & 0.7967 \\ 
    \ \ +\textit{Language Steering} (De.) & \textbf{0.7509} & \textbf{0.7605} & \textbf{0.7808}$^{*}$ & 0.8200 & \textbf{0.8746} & \textbf{0.8916}$^{***}$ & \textbf{0.8131}$^{*}$ \\ 

    \bottomrule
    \end{tabular}
    \caption{Culture representation scores for China and Germany under different steering methods for LLMs. 
    The content in ``()'' is to indicate the language being used, where ``En.'' denotes English, ``Zh.'' denotes Chinese, and ``De.'' denotes German. In all cases, the language of our inputs (task instruction, few-shot examples, and question) always keeps the same.
    Moreover, the significance is assessed using the $t$-test: * denotes $p$-value $<$ 0.05, ** denotes $p$-value $<$ 0.01, and *** denotes $p$-value $<$ 0.001.}
    \label{tab:slm_steerability_results}
\end{table*}

\subsection{Other Matrics}
\label{sec:g5}

In addition to \textit{alignment}, previous work also employed the normalized mean absolute error (NMAE) and Jensen-Shannon divergence (JSD) to calculate the distance between distributions~\citep{liu2024robust}.
In this section, we employ NMAE and JSD as alignment scores to validate Pearson's $r$ between the current metric $alignment$ and both NMAE and JSD.
As shown in Table~\ref{tab:other_matrics_aya}-\ref{tab:other_matrics_dsr1}, the three alignment metrics \textit{alignment}, NMAE, and JSD keep high Pearson's $r$ across different models, indicating consistency among the alignment metrics.

\begin{table}[h]
    \centering
    \begin{tabular}{lccc}
            \textbf{} &         \textbf{Alignment} & \textbf{NMAE}  & \textbf{JSD}  \\
            \midrule
            \textbf{Alignment}  & 1                &  0.9150        & 0.8757        \\
            \textbf{NMAE}       &                  &  1             & 0.9390        \\
            \textbf{JS div}     &                  &                & 1             \\
    \end{tabular}
    \caption{Pearson's $r$ between different matrics on Aya-23-35B.}
    \label{tab:other_matrics_aya}
\end{table}

\begin{table}[h]
    \centering
    \begin{tabular}{lccc}
            \textbf{} &         \textbf{Alignment} & \textbf{NMAE}  & \textbf{JSD}  \\
            \midrule
            \textbf{Alignment}  & 1                &  0.8696        & 0.8584        \\
            \textbf{NMAE}       &                  &  1             & 0.8920        \\
            \textbf{JS div}     &                  &                & 1             \\
    \end{tabular}
    \caption{Pearson's $r$ between different matrics on Llama3-70B-Instruct.}
    \label{tab:other_matrics_llama}
\end{table}

\begin{table}[h]
    \centering
    \begin{tabular}{lccc}
            \textbf{} &         \textbf{Alignment} & \textbf{NMAE}  & \textbf{JSD}  \\
            \midrule
            \textbf{Alignment}  & 1                &  0.9102        & 0.9040        \\
            \textbf{NMAE}       &                  &  1             & 0.8921       \\
            \textbf{JS div}     &                  &                & 1             \\
    \end{tabular}
    \caption{Pearson's $r$ between different matrics on Qwen2.5-72B-Instruct.}
    \label{tab:other_matrics_qwen}
\end{table}

\begin{table}[h]
    \centering
    \begin{tabular}{lccc}
            \textbf{} &         \textbf{Alignment} & \textbf{NMAE}  & \textbf{JSD}  \\
            \midrule
            \textbf{Alignment}  & 1                &  0.9287        & 0.9259        \\
            \textbf{NMAE}       &                  &  1             & 0.8751      \\
            \textbf{JS div}     &                  &                & 1             \\
    \end{tabular}
    \caption{Pearson's $r$ between different matrics on GPT-3.5-Turbo.}
    \label{tab:other_matrics_gpt35}
\end{table}

\begin{table}[h]
    \centering
    \begin{tabular}{lccc}
            \textbf{} &         \textbf{Alignment} & \textbf{NMAE}  & \textbf{JSD}  \\
            \midrule
            \textbf{Alignment}  & 1                &  0.9125        & 0.9113       \\
            \textbf{NMAE}       &                  &  1             & 0.9268      \\
            \textbf{JS div}     &                  &                & 1             \\
    \end{tabular}
    \caption{Pearson's $r$ between different matrics on GPT-4.}
    \label{tab:other_matrics_gpt4}
\end{table}

\begin{table}[h]
    \centering
    \begin{tabular}{lccc}
            \textbf{} &         \textbf{Alignment} & \textbf{NMAE}  & \textbf{JSD}  \\
            \midrule
            \textbf{Alignment}  & 1                &  0.9244       & 0.9294     \\
            \textbf{NMAE}       &                  &  1             & 0.9362      \\
            \textbf{JS div}     &                  &                & 1             \\
    \end{tabular}
    \caption{Pearson's $r$ between different matrics on GPT-5.}
    \label{tab:other_matrics_gpt5}
\end{table}

\begin{table}[h]
    \centering
    \begin{tabular}{lccc}
            \textbf{} &         \textbf{Alignment} & \textbf{NMAE}  & \textbf{JSD}  \\
            \midrule
            \textbf{Alignment}  & 1                &  0.9094      & 0.9195    \\
            \textbf{NMAE}       &                  &  1             & 0.8975     \\
            \textbf{JS div}     &                  &                & 1             \\
    \end{tabular}
    \caption{Pearson's $r$ between different matrics on DeepSeek-V3.}
    \label{tab:other_matrics_dsv3}
\end{table}

\begin{table}[h]
    \centering
    \begin{tabular}{lccc}
            \textbf{} &   \textbf{Alignment} & \textbf{NMAE} & \textbf{JSD} \\
            \midrule
            \textbf{Alignment} & 1 &  0.8898 & 0.9245\\
            \textbf{NMAE} & & 1 & 0.9004 \\
            \textbf{JS div} & & & 1 \\
    \end{tabular}
    \caption{Pearson's $r$ between different matrics on DeepSeek-R1.}
    \label{tab:other_matrics_dsr1}
\end{table}

\subsection{Other Baselines}
\label{sec:g6}

\begin{figure}[h]
\centering
\includegraphics[width=\linewidth]{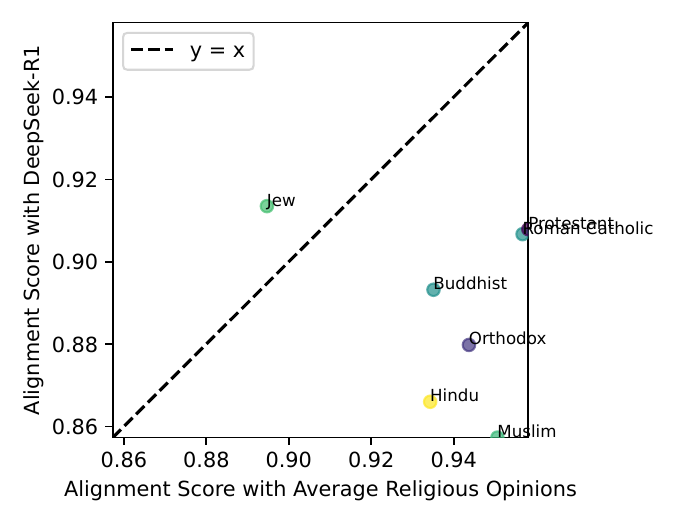}
\caption{Religion as the baseline.}
\label{fig:religion_as_the_baseline}
\end{figure}

\begin{figure}[h]
\centering
\includegraphics[width=\linewidth]{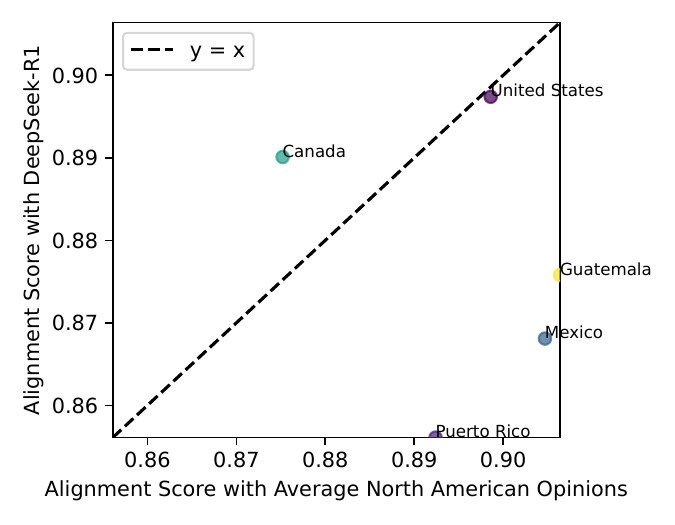}
\caption{Region (America) as the baseline.}
\label{fig:region_am_as_the_baseline}
\end{figure}

\begin{figure}[h]
\centering
\includegraphics[width=\linewidth]{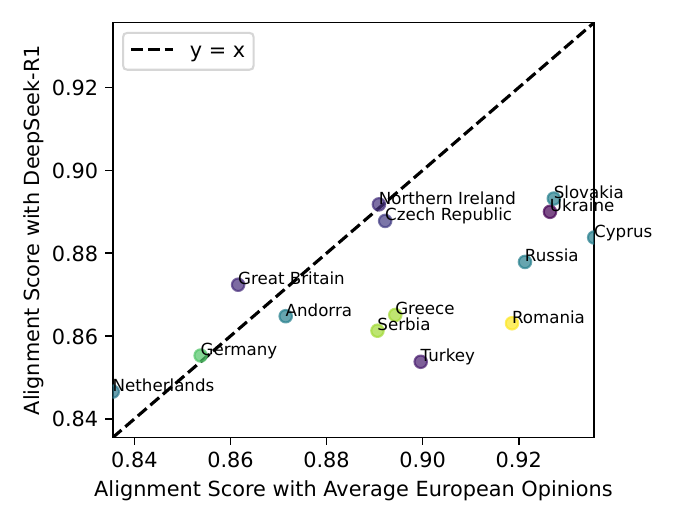}
\caption{Region (Europe) as the baseline.}
\label{fig:region_eu_as_the_baseline}
\end{figure}

Take DeepSeek-R1 as an example, we also perform a religious analysis, using religion as a baseline. 
As shown in Figure~\ref{fig:religion_as_the_baseline}, most religions are under-aligned except Jew, which is appropriately aligned.
In addition, as shown in Figures~\ref{fig:region_am_as_the_baseline} and \ref{fig:region_eu_as_the_baseline}, we include regional (e.g., North American and Europe) averages to enrich our baseline.
For example, within the North American region, Canada is over-aligned, the US is appropriately aligned, while Mexico, Puerto Rico, and Guatemala are under-aligned.

\section{Limitations}
We acknowledge that our study has limitations, one of which is that it is limited by the scope of the WVS and cannot cover all countries.
The second limitation is that, due to human resource constraints, we cannot examine all languages supported by the WVS. However, we are fortunate to have covered eight languages (Spanish, Chinese, Japanese, Korean, German, Russian, Vietnamese, and Portuguese) and report exciting findings. Another limitation is that we did not consider whether there were cross-linguistic effects if a country conducted its survey using a multilingual WVS questionnaire. At the very least, we hope that these limitations can serve as directions for future studies.

\end{document}